\documentclass{article}
\usepackage[utf8]{inputenc}
\usepackage[a4paper, margin=2.5cm]{geometry}
\usepackage{amsmath}
\usepackage{setspace}
\usepackage{placeins}
\usepackage{booktabs}
\usepackage{siunitx}
\usepackage{makecell}
\usepackage{dblfloatfix}
\usepackage{svg}
\usepackage{graphicx}
\usepackage{float}

\title{A hybrid analytical-PINN model for subsurface simulation of geothermal heat exchangers in heterogeneous underground}
\author{
Moke Rao, Thomas Hamacher, Smajil Halilovic\thanks{Corresponding author. Email: smajil.halilovic@tum.de}\\[0.5em]
\small\textit{Technical University of Munich, Chair of Renewable and Sustainable Energy Systems, Germany}
}
\date{}

\usepackage{graphicx}
\usepackage{subcaption}
\usepackage{booktabs}
\usepackage{array}
\usepackage{booktabs}
\usepackage{multirow}
\usepackage{caption}
\usepackage[numbers,sort&compress]{natbib}

\begin{document}

\maketitle

\noindent\rule{\textwidth}{0.4pt}

\vspace{0.8em}
\noindent{\normalfont\Large\bfseries Abstract\par}
\vspace{0.8em}
\noindent
Accurate and efficient prediction of subsurface temperature fields is essential for the design and operation of borehole heat exchanger (BHE) systems. Here we develop a parametric hybrid analytical and physics-informed neural network (PINN) framework for long-term multi-BHE simulations in heterogeneous underground. The method analytically extracts the singular line source response and enables the effective training of neural correction associated with subsurface heterogeneity. An explicit parametrization of the thermal conductivity allows physics-informed learning of a single feedforward neural network to generalize across different subsurface conditions. By formulating the correction in borehole-centered relative coordinates, the learned correction can be reused as a universal corrector through spatial and temporal superposition principles. Numerical experiments based on the infinite line source (ILS), finite line source (FLS) and moving finite line source (MFLS) models show that the hybrid method outperforms analytical approximations with stable accuracy over long simulation horizons and achieves orders-of-magnitude speedups over traditional solvers. The proposed framework therefore combines the efficiency of analytical models with the ability of numerical methods to capture heterogeneous subsurface physics, providing a fast and accurate approach for repeated long-term simulation of multi-BHE systems.

\vspace{0.6em}

\vspace{0.8em}
\noindent\rule{\textwidth}{0.4pt}

\vspace{0.5em}

\section{Introduction}
Decarbonizing heating and cooling is a central component of the transition toward low-carbon energy systems \cite{figueira2024shallow, halilovic2022integration}. Shallow geothermal energy, commonly utilized through ground-source heat-pump systems, provides a locally available and seasonally stable thermal resource for building thermal conditioning and underground thermal-energy storage \cite{sadeghi2024review, halilovic2024spatial, romanov2022geothermal}. Among the available ground heat exchanger configurations, vertical borehole heat exchangers (BHE) are particularly attractive. A BHE circulates a heat-carrier fluid through closed pipes installed in a vertical borehole and exchanges heat with the surrounding ground without directly extracting groundwater \cite{randow2022modeling}. Its long-term performance depends strongly on the transient thermal response of the subsurface and on the thermal interaction among neighboring boreholes \cite{cassina2022thermal, heinzel2026optimization}.

Accurate and repeated forward simulation of the ground temperature field is therefore essential for the design and operation of BHE systems \cite{zhang2018computational,dePaly2012Optimization}. A broad range of analytical and semi-analytical models was developed for BHE simulation, including infinite line source, finite line source and moving finite line source models \cite{dePaly2012Optimization, Zeng2002FLS, MolinaGiraldo2011MFLS}. They represent a borehole as a line source and provide physically interpretable descriptions of the radial transient response. However, their computation efficiency is obtained at the expense of restrictive assumptions, such as idealized homogeneous and infinite ground domain, simplified boundary conditions and spatially uniform groundwater velocity \cite{halilovic2023lahm}. On the other hand, numerical methods provide greater flexibility and can resolve more complex soil properties, geometries and subsurface transport conditions \cite{alkhoury2010efficient, li2015review, halilovic2022well}. Over the last decades, a variety of specialized numerical solvers for BHE simulation have been developed, including TRNSYS \cite{hellstrom1989duct}, OpenGeoSys \cite{shao2016ogs} and FEFLOW \cite{diersch2011bhe1}. However, long-term three-dimensional numerical simulations remain computationally prohibitive, especially for repeated evaluations under varying configurations and operating conditions \cite{halilovic2023optimization}.

Applying deep neural networks (DNNs) to repeated BHE and subsurface thermal simulation has thus drawn attention recently. A trained neural surrogate can predict temperature fields at a substantially lower online cost compared to numerical simulation and especially benefits any procedure requiring numerous simulation runs. In terms of training mechanism, the data-driven and physics-informed neural networks define the loss based on different sources of information. The quality and quantity of training samples largely determine the performance of the purely data-driven models. In contrast, physics-informed neural networks augment the learning process by introducing governing partial differential equations (PDEs), initial conditions and boundary conditions into the training objective \cite{raissi2019physics}. Consequently, the use of physical constraints can reduce the expensive data generation burden, allow differentiable approximations over the space-time domain and enhance the generalization ability outside the sampled parameter range. These properties make PINNs attractive as reusable thermal simulators for BHE systems in heterogeneous media. However, their utilization in the field is still lacking due to the challenging nature of the singular (BHE) sources \cite{halilovic2026deep}.      

Solving PDEs with singular sources is challenging for standard DNNs and vanilla PINNs since they lack the expressive power to effectively approximate the solution containing a highly localized singularity. Therefore, several developments for physics-informed neural solvers have been proposed to deal with the nonsmoothness of PDE solution. Gao et al. \cite{gao2023failureinformed} proposed a failure-informed adaptive sampling strategy that places additional collocation points in regions with large residual-based failure probabilities. This is further combined with re-sampling and subset simulation to improve the detection of localized high-error regions in \cite{gao2024failureinformedpart2}. Adaptive sampling is also adapted in \cite{Kara2025DeepONetFEM} where the localized source is represented by a normalized Gaussian profile. For the approximation of point sources, Huang et al. \cite{huang2022universal} use a continuous probability density and combine this regularization with an uncertainty-based loss weighting strategy and a practical multiscale neural architecture. Making use of the separation of the singularity from the network approximation target, \citet{hu2023singularitysplitting} proposed a singularity-splitting deep Ritz method that represents the known singular component analytically and trains the neural network only for the regular remainder. Furthermore, Hu et al. \cite{hu2024sepinn} developed a singularity-enriched PINN that explicitly incorporates explicit singular basic functions for Poisson problems to construct a more expressive approximation space of PINN. Later this framework was extended to solving inverse problems by jointly identifying the locations and intensities of unknown point sources. \citet{lai2022shallowritz} handle elliptic problems with singular sources on an interface by formulating the variational problem and use level set function to give the network additional information about the interface.      

In this paper, we aim to enable fast and accurate simulation of BHE-induced heat transport in heterogeneous media by combining the computational efficiency of analytical models with the ability of numerical methods to capture heterogeneous subsurface physics. To this end, we propose a parametric hybrid analytical–PINN framework that decomposes the temperature response into an analytical base solution and a physics-informed learned correction. The contributions of the proposed method are three-fold. First, we exclude the delta function singularity embedded in the source terms by using analytical solutions as base models. Second, we target the linearly changing thermal conductivity cases, which allows training the network with physics-informed loss and generalization to unseen gradient conductivity scenarios. Third, the parametric neural network is trained on various sampled conductivity conditions to approximate the temperature correction induced by a singular borehole at the origin with unit heat transfer rate, which could be employed as a computationally efficient universal corrector by applying superposition principle. This helps avoid training different networks for diverse arrangements of BHEs and source load profiles. In addition, we rewrite the governing equation into a formulation on relative coordinates with respect to the borehole location, so that the network learns the temperature correction as a function of the source-relative position under prescribed thermal conductivity and groundwater flow conditions in an open-boundary domain. We thus include the distance from the borehole as input feature of the network that effectively improves the simulation accuracy near the source. To the best of our knowledge, this work represents the first application of a hybrid analytical–PINN decomposition framework to subsurface thermal simulation of BHEs.         

The rest of the paper is organized as follows. Section \ref{Problem_formulation} introduces the classical formulations for the global soil temperature model problem. In Section \ref{methodology}, the parametric hybrid-PINN universal corrector method to simulate the models is presented. In Section \ref{case_study}, test cases are presented to evaluate the accuracy and computational efficiency of the proposed ILS, FLS and MFLS based models against numerical simulations, followed by the conclusions and future work directions in Section \ref{conclusion}.

\section{Problem Formulation}\label{Problem_formulation}

Looking into the transient propagation of temperature perturbations in the subsurface induced by a field of BHEs, we consider the global thermal problem as the evolution of the soil temperature field around the BHEs. Assume $\Omega_s$ denotes the soil domain and $I=\left(0,T\right)$ is the time interval of interest. The soil temperature $T\left(\mathbf{x},t\right)$, $\mathbf{x}\in \Omega_s$, $t\in I$ in a heterogeneous porous medium is governed by the combined effects of thermal diffusion and groundwater advection. We first consider the conduction-dominated case \cite{Zeng2002FLS}
\begin{align}
    \rho_s C_s\frac{\partial T}{\partial t}
    -\nabla \cdot \left( \lambda(\mathbf{x}) \nabla T \right)
    =f\left(\mathbf{x},t\right),\label{eq:conduction_soil_model}
\end{align}
where $\rho_s C_s$ is the volumetric heat capacity of the soil and $\lambda(\mathbf{x})$ denotes the soil thermal conductivity. Hence the temperature change $\widetilde{T}\left(\mathbf{x},t\right)=T\left(\mathbf{x},t\right)-T_u$ relative to the undisturbed constant ground state $T_{u}$ follows the same mechanism
\begin{align}
    \rho_s C_s\frac{\partial \widetilde{T}}{\partial t}
    -\nabla \cdot \left( \lambda(\mathbf{x}) \nabla \widetilde{T} \right)
    =f\left(\mathbf{x},t\right).\label{eq:conduction_soil_change_model}
\end{align}
Taking the groundwater flow effect into account, 
BHE-induced temperature change satisfies the advection-diffusion equation \cite{diao2004heat}
\begin{equation}
\rho C \frac{\partial \widetilde{T}}{\partial t}
+
\rho_fC_f \mathbf{v}\cdot \nabla \widetilde{T}
-
\nabla \cdot \left( \lambda(\mathbf{x}) \nabla \widetilde{T} \right)
=
f(\mathbf{x},t),
\label{eq:advection_conduction_soil_model}
\end{equation}
where $\rho C$ is the combined volumetric heat capacity of the solid and liquid phase, i.e., $\rho C=\epsilon \rho_f C_f +\left(1-\epsilon\right)\rho_s C_s$, $\rho_f C_f$ is the volumetric heat capacity of groundwater, $\epsilon$ is the porosity and $\mathbf{v}$ is the Darcy velocity of the groundwater flow.

Furthermore, $f\left(\mathbf{x},t\right)$ represents the heat source or sink induced by BHE operation. For a field of $K$ vertical boreholes, the heat source term can be written as a superposition of individual borehole sources,
\begin{equation}
f(\mathbf{x},t)
=
\sum_{k=1}^{K} f_k(\mathbf{x},t),
\notag
\end{equation}
where $f_k$ denotes the source term associated with the $k$-th borehole. Let $\mathbf{x}^k=(x_{\mathrm{BHE}}^k,y_{\mathrm{BHE}}^k)$ be the location of the $k$-th borehole in the horizontal plane and $q_k(t)$ be the heat injection or extraction rate per unit length. The infinite line source idealizes the borehole as an infinitely long vertical source and writes
\begin{equation}
f_k^{\mathrm{ILS}}(\mathbf{x},t)
=
q_k(t)\,
\delta(x-x_{\mathrm{BHE}}^k)\delta(y-y_{\mathrm{BHE}}^k),
\label{eq:ils_source}
\end{equation}
where $\delta(\cdot)$ is the Dirac delta distribution.

A more realistic representation accounts for the finite borehole length $L$. If the borehole extends from the depth $z=D$ to the depth $z=D+L$, the finite line source term reads
\begin{equation}
f_k^{\mathrm{FLS}}(\mathbf{x},t)
=
q_k(t)\,
\delta(x-x_{\mathrm{BHE}}^k)\delta(y-y_{\mathrm{BHE}}^k)
\chi_{[D,D+L]}(z),
\label{eq:fls_source}
\end{equation}
where $\chi_{[D,D+L]}(z)$ is the indicator function along the borehole depth. 

The temperature field is initialized from the undisturbed state 
\begin{equation}
T(\mathbf{x},0)=T_u,\quad \mathbf{x}\in\Omega_s.
\label{total_temperature_initial_condition}
\end{equation}
Equivalently, for the temperature change $\widetilde{T}$, this condition is enforced as
\begin{equation}
\widetilde{T}(\mathbf{x},0)=0,\quad \mathbf{x}\in\Omega_s.\label{eq:initial_Ttilde}
\end{equation} 

In this work, we focus on a heterogeneous setting in which the soil thermal conductivity is modeled as a linear gradient field in the horizontal plane
\begin{equation}
\lambda(\mathbf{x})
=
\lambda_c+\lambda_xx+\lambda_yy,
\notag
\end{equation}
where $\lambda_c$ denotes the conductivity value at the horizontal coordinate origin $\left(x,y\right)=\left(0,0\right)$ and $\left(\lambda_x,\lambda_y\right)$ is the constant gradient of spatially varying $\lambda\left(\mathbf{x}\right)$. The conductivity is assumed to be invariant along the vertical direction. Fully resolved subsurface conductivity fields are rarely available in practical shallow geothermal applications, while standard thermal response tests usually provide only effective or averaged thermal properties. Therefore, the gradient conductivity field is adopted as a controlled representation of commonly encountered BHE system-scale subsurface heterogeneity. Compared with the homogeneous assumption, it captures more realistic subsurface heat transfer characteristics while remaining sufficiently simple for systematic analysis. 

\section{Hybrid-PINN Universal Corrector Method}\label{methodology}
\subsection{Hybrid PINN Corrector Based on Analytical Models}

We propose a hybrid PINN corrector method for problems (\ref{eq:conduction_soil_change_model}) and (\ref{eq:advection_conduction_soil_model}) subject to condition (\ref{eq:initial_Ttilde}) by applying analytical models with respect to source terms (\ref{eq:ils_source}) and (\ref{eq:fls_source}). In this section, we discuss the case with a single source located in $\Omega_s$ with a unity load rate. The case of multiple boreholes can be handled via spatial superposition. 

Instead of approximating the entire BHE-induced temperature field, we decompose the temperature change into an analytical base component and a learned correction component
\begin{equation}
\widetilde{T}(\mathbf{x},t)
=
\widetilde{T}^{m}(\mathbf{x},t)
+
\widetilde{T}_c^{m}(\mathbf{x},t),
\qquad
m\in\{\mathrm{ILS},\mathrm{FLS},\mathrm{MFLS}\}.
\notag
\end{equation}
Here, $\widetilde{T}^{m}$ denotes the analytical prediction given by the selected line source model, and $\widetilde{T}_C^{m}$ is the correction field to be approximated by a neural network. By incorporating analytical knowledge of the underlying solution to the problem of interest, the singularity due to the line source could be captured explicitly. The correction component is then introduced to represent the discrepancy between the analytical reference approximation and global heterogeneous solution. We consider three analytical base models depending on whether the groundwater flow and finite borehole length are taken into account. 
\paragraph{Infinite line source (ILS) model.}  For a two-dimensional infinite homogeneous domain, ILS model assumes heat conduction without groundwater advection in the underground and expresses the radial temperature distribution caused by a single BHE subjected to a piecewise constant heat load over $N_t$ time steps as
\begin{equation}
\widetilde{T}^{\mathrm{ILS}}(x,y,t)
=
\sum_{l=0}^{N_t-1}
\frac{q_{l}-q_{l-1}}{4\pi\lambda_0}
\mathrm{E_1}\!\left(
\frac{r^2}{4\alpha_0\left(t-t_{l}\right)}
\right),
\label{eq:ils_piecewise_load}
\end{equation}
where $\mathrm{E_1}\left(\cdot\right)$ is the exponential integral, $r=\sqrt{(x-x_{\mathrm{BHE}})^2+(y-y_{\mathrm{BHE}})^2}$ is the distance from the BHE centered at $\left(x_{\mathrm{BHE}},y_{\mathrm{BHE}}\right)$, $\lambda_0$ is the bulk thermal conductivity, $\alpha_0=\lambda_0/\left(\rho_s C_s\right)$ denotes the thermal diffusivity and $q_l$ is the load during time step from $t_{l}$ to $t_{l+1}$, with $q_{-1}=0$ and $t_0=0$ \cite{dePaly2012Optimization}. The corresponding homogeneous heat diffusion equation with respect to ILS solution is
\begin{equation}
\rho_s C_s
\frac{\partial \widetilde{T}^{\mathrm{ILS}}}{\partial t}
-
\lambda_0
\Delta \widetilde{T}^{\mathrm{ILS}}
=q(t)\delta\left(x-x_{\mathrm{BHE}}\right)\delta\left(y-y_{\mathrm{BHE}}\right).
\label{eq:ils_homogeneous_pde}
\end{equation}

\paragraph{Finite line source (FLS) model.}
To evaluate the thermal behavior in a three-dimensional homogeneous domain, FLS model represents the boreholes as line sources of finite lengths and assumes a uniform heat rate per unit length along the borehole, an idealized geometry with zero radius and no groundwater effect on the temperature distribution in the underground \cite{Zeng2002FLS}. If the borehole extends from the ground surface $z=D$ to the depth $z=D+L$, 
\begin{equation}
\widetilde{T}^{\mathrm{FLS}}(x,y,z,t)
=
\sum_{l=0}^{N_t-1}
\frac{q_l-q_{l-1}}{4\pi \lambda_0}
\int_{D}^{D+L}
\frac{
\operatorname{erfc}
\!\left[
\dfrac{R(z')}{2\sqrt{\alpha_0\left(t-t_l\right)}}
\right]
}
{R(z')}
\,\mathrm dz',
\label{eq:fls_piecewise_load}
\end{equation}
where $\mathrm{erfc}\left(\cdot\right)$ denotes the complementary error function and $R(z')=\sqrt{r^2+(z-z')^2}$. The corresponding homogeneous equation regarding FLS solution is 
\begin{equation}
\rho_s C_s
\frac{\partial \widetilde{T}^{\mathrm{FLS}}}{\partial t}
-
\lambda_0
\Delta \widetilde{T}^{\mathrm{FLS}}
=q(t)
\delta(x-x_{\mathrm{BHE}})
\delta(y-y_{\mathrm{BHE}})
\chi_{[D,D+L]}(z).
\notag
\end{equation}
\paragraph{Moving finite line source (MFLS) model.}
The MFLS model extends the FLS solution to a homogeneous medium with groundwater advection governed by a uniform Darcy velocity $\mathbf{v}=\left(v_x,v_y,0\right)$, which reads
\begin{equation}
\widetilde{T}^{\mathrm{MFLS}}(x,y,z,t)
=
\sum_{l=0}^{N_t-1}
\frac{q_l-q_{l-1}}{8\pi \lambda_0}
\int_{0}^{t-t_l}
\frac{1}{\tau}
\exp
\!\left[
-
\frac{
(x-\frac{\rho_fC_f}{\rho C}v_x\tau)^2+(y-\frac{\rho_fC_f}{\rho C}v_y\tau)^2
}
{4\alpha'_0\tau}
\right]
\mathcal{Z}(z,\tau)
\,\mathrm d\tau ,
\label{eq:mfls_piecewise_load}
\end{equation}
where $\alpha'_0=\lambda_0/\left(\rho C\right)$,
\begin{equation}
\mathcal{Z}(z,\tau)
=
\operatorname{erf}
\!\left[
\frac{D+L-z}{2\sqrt{\alpha'_0\tau}}
\right]
-
\operatorname{erf}
\!\left[
\frac{D-z}{2\sqrt{\alpha'_0\tau}}
\right],
\notag
\end{equation}
and $\mathrm{erf}\left(\cdot\right)$ denotes the error function \cite{MolinaGiraldo2011MFLS}. The corresponding homogeneous advection-diffusion equation concerning MFLS solution is
\begin{equation}
\rho C
\frac{\partial \widetilde{T}^{\mathrm{MFLS}}}{\partial t}
+
\rho_f C_f \mathbf{v}\cdot
\nabla \widetilde{T}^{\mathrm{MFLS}}
-
\lambda_0
\Delta \widetilde{T}^{\mathrm{MFLS}}=q(t)
\delta(x-x_{\mathrm{BHE}})
\delta(y-y_{\mathrm{BHE}})
\chi_{[D,D+L]}(z).
\notag
\end{equation}

All three analytical models are used here as reference solutions under simplified homogeneous assumptions. They play a key role in removing the dominant singular response so that the remaining correction field can be learned more efficiently by the physics-informed neural network. We now derive the governing equations for $\widetilde{T}_c^m$, $m=\{\mathrm{ILS}, \mathrm{FLS}, \mathrm{MFLS}\}$. We illustrate the procedure using $\widetilde{T}_c^{\mathrm{ILS}}$ as an example, and the deduction for $\widetilde{T}_c^{\mathrm{FLS}}$ and $\widetilde{T}_c^{\mathrm{MFLS}}$ can be handled similarly. 

Setting $f\left(\mathbf{x},t\right)=q(t)\delta\left(x-x_{\mathrm{BHE}}\right)\delta\left(y-y_{\mathrm{BHE}}\right)$ in (\ref{eq:conduction_soil_change_model}) and subtracting (\ref{eq:ils_homogeneous_pde}) from it, we have the correction equation
\begin{equation}
\rho_s C_s\frac{\partial \widetilde{T}_{c}^{\mathrm{ILS}}}{\partial t}-\nabla \cdot\left(\lambda\left(x,y\right)\nabla \widetilde{T}_{c}^{\mathrm{ILS}}\right)=\nabla \cdot \left(\left(\lambda\left(x,y\right)-\lambda_0\right)\nabla \widetilde{T}^{\mathrm{ILS}}\right),\label{correction_ils}
\end{equation}
where the left-hand side describes the transport of temperature correction and the forcing term represents the heat flow error introduced by plugging the homogeneous solution into the heterogeneous physics $\lambda\left(\mathbf{x}\right)$.
By rewriting $\lambda\left(\mathbf{x}\right)$ in (\ref{correction_ils}) as $\lambda\left(\mathbf{x}\right)=\lambda_B+\lambda_x x_{\mathrm{rel}}+\lambda_y y_{\mathrm{rel}}$, where $\lambda_B=\lambda\left(x_{\mathrm{BHE}},y_{\mathrm{BHE}}\right)$, $x_{\mathrm{rel}}=x-x_{\mathrm{BHE}}$ and $y_{\mathrm{rel}}=y-y_{\mathrm{BHE}}$, we obtain
\begin{align}
& \rho_s C_s\frac{\partial \widetilde{T}_{c}^{\mathrm{ILS}}}{\partial t}\left(x,y\right)
- \lambda_x\frac{\partial \widetilde{T}_{c}^{\mathrm{ILS}}}{\partial x}\left(x,y\right)
-\left( \lambda_B + \lambda_x x_{\mathrm{rel}} +\lambda_y y_{\mathrm{rel}}\right) \frac{\partial^2 \widetilde{T}_{c}^{\mathrm{ILS}}}{\partial x^2}\left(x,y\right)\notag\\
&- \lambda_y\frac{\partial \widetilde{T}_{c}^{\mathrm{ILS}}}{\partial y}\left(x,y\right)
-\left( \lambda_B + \lambda_x x_{\mathrm{rel}} +\lambda_y y_{\mathrm{rel}}\right) \frac{\partial^2 \widetilde{T}_{c}^{\mathrm{ILS}}}{\partial y^2}\left(x,y\right)\notag\\
= &  \lambda_x\frac{\partial \widetilde{T}^{\mathrm{ILS}}}{\partial x}\left(x_{\mathrm{rel}},y_{\mathrm{rel}}\right)
+ \left(\lambda_B-\lambda_0 + \lambda_x x_{\mathrm{rel}} +\lambda_y y_{\mathrm{rel}}\right) \frac{\partial^2 \widetilde{T}^{\mathrm{ILS}}}{\partial x^2}\left(x_{\mathrm{rel}},y_{\mathrm{rel}}\right)\notag\\
&+ \lambda_y\frac{\partial \widetilde{T}^{\mathrm{ILS}}}{\partial y}\left(x_{\mathrm{rel}},y_{\mathrm{rel}}\right)
+ \left(\lambda_B-\lambda_0 + \lambda_x x_{\mathrm{rel}} +\lambda_y y_{\mathrm{rel}}\right)\frac{\partial^2 \widetilde{T}^{\mathrm{ILS}}}{\partial y^2}\left(x_{\mathrm{rel}},y_{\mathrm{rel}}\right).\label{showing_coordinates_absolute}
\end{align}
Replacing $x$ by $x_{\mathrm{rel}}+x_{\mathrm{BHE}}$ and $y$ by $y_{\mathrm{rel}}+y_{\mathrm{BHE}}$ in (\ref{showing_coordinates_absolute}) and putting the horizontal coordinate origin at the central location of the borehole result in
\begin{align}
& \rho_s C_s\frac{\partial \widetilde{T}_{c}^{\mathrm{ILS}}}{\partial t}\left(x_{\mathrm{rel}},y_{\mathrm{rel}}\right)
- \lambda_x\frac{\partial \widetilde{T}_{c}^{\mathrm{ILS}}}{\partial x}\left(x_{\mathrm{rel}},y_{\mathrm{rel}}\right)
-\left( \lambda_B + \lambda_x x_{\mathrm{rel}} +\lambda_yy_{\mathrm{rel}}\right) \frac{\partial^2 \widetilde{T}_{c}^{\mathrm{ILS}}}{\partial x^2}\left(x_{\mathrm{rel}},y_{\mathrm{rel}}\right)\notag\\
&- \lambda_y\frac{\partial \widetilde{T}_{c}^{\mathrm{ILS}}}{\partial y}\left(x_{\mathrm{rel}},y_{\mathrm{rel}}\right)
- \left( \lambda_B + \lambda_x x_{\mathrm{rel}} +\lambda_yy_{\mathrm{rel}}\right) \frac{\partial^2 \widetilde{T}_{c}^{\mathrm{ILS}}}{\partial y^2}\left(x_{\mathrm{rel}},y_{\mathrm{rel}}\right)\notag\\
= &  \lambda_x\frac{\partial \widetilde{T}^{\mathrm{ILS}}}{\partial x}\left(x_{\mathrm{rel}},y_{\mathrm{rel}}\right)
+ \left(\lambda_B-\lambda_0 + \lambda_x x_{\mathrm{rel}} +\lambda_y y_{\mathrm{rel}}\right) \frac{\partial^2 \widetilde{T}^{\mathrm{ILS}}}{\partial x^2}\left(x_{\mathrm{rel}},y_{\mathrm{rel}}\right)\notag\\
&+ \lambda_y\frac{\partial \widetilde{T}^{\mathrm{ILS}}}{\partial y}\left(x_{\mathrm{rel}},y_{\mathrm{rel}}\right)
+ \left( \lambda_B -\lambda_0+ \lambda_x x_{\mathrm{rel}} +\lambda_y y_{\mathrm{rel}}\right) \frac{\partial^2 \widetilde{T}^{\mathrm{ILS}}}{\partial y^2}\left(x_{\mathrm{rel}},y_{\mathrm{rel}}\right).\label{showing_coordinates_relative}
\end{align}
Thus, the solution of (\ref{showing_coordinates_relative}) represents the temperature correction distribution as a function of the relative position with respect to the borehole. 

Assuming FLS reference model uses the same source term as the heterogeneous problem and applying a similar procedure to obtain the correction equation for FLS base solution arrives at
\begin{equation}
\rho_s C_s
\frac{\partial \widetilde{T}_{c}^{\mathrm{FLS}}}{\partial t}
-
\nabla\cdot
\left(
\lambda(x,y)\nabla \widetilde{T}_{c}^{\mathrm{FLS}}
\right)
=
\nabla\cdot
\left(
\left(\lambda(x,y)-\lambda_0\right)
\nabla \widetilde{T}^{\mathrm{FLS}}
\right).
\label{eq:correction_fls}
\end{equation}
Here, the gradient and divergence operators are taken in three-dimensional space. The same idea leads to the MFLS-based correction equation
\begin{equation}
\rho C
\frac{\partial \widetilde{T}_{c}^{\mathrm{MFLS}}}{\partial t}
+
\rho_f C_f \mathbf{v}\cdot
\nabla \widetilde{T}_{c}^{\mathrm{MFLS}}
-
\nabla\cdot
\left(
\lambda(x,y)\nabla \widetilde{T}_{c}^{\mathrm{MFLS}}
\right)
=
\nabla\cdot
\left(
\left(\lambda(x,y)-\lambda_0\right)
\nabla \widetilde{T}^{\mathrm{MFLS}}
\right).
\label{eq:correction_mfls}
\end{equation}
Via formulating $\lambda\left(\mathbf{x}\right)$ as $\lambda_B+\lambda_xx_{\mathrm{rel}}+\lambda_yy_{\mathrm{rel}}$ in (\ref{eq:correction_fls}) and (\ref{eq:correction_mfls}), the correction equation based on an arbitrary analytical base model describes the correction field in source-relative coordinates, enabling the construction of a reusable universal corrector.  

As a result, the ILS, FLS and MFLS corrections embedded in the differential equations obey the same mechanism, where the left-hand side of the equations transport the thermal correction under the target global physics, while the right-hand side provides the residual forcing generated by evaluating a homogeneous analytical prior in a heterogeneous conductivity field. Thus, the neural network does not need to directly deal with the singularity challenge arising from the original problems (\ref{eq:conduction_soil_change_model}) and (\ref{eq:advection_conduction_soil_model}).

Since the temperature perturbation satisfies (\ref{eq:initial_Ttilde})
and the hybrid decomposition gives $\widetilde{T}_{c}^{m}=\widetilde{T}-\widetilde{T}^{m}$, $m\in\{\mathrm{ILS},\mathrm{FLS},\mathrm{MFLS}\}$, the corresponding correction field satisfies the following initial condition
\begin{equation}
\widetilde{T}_{c}^{m}(\mathbf{x},0)=0, \quad \mathbf{x}\in \Omega_s, 
 \quad \mathrm{m}\in\{\mathrm{ILS}, \mathrm{FLS}, \mathrm{MFLS}\}.\notag
\end{equation}

\subsection{A Parametric PINN Method}\label{parametric_pinn_method}
To avoid retraining a separate network for each borehole location and each thermal conductivity realization, we formulate the neural corrector as a parametric PINN. The model takes the local conductivity parameters and the
spatiotemporal coordinates as inputs, and therefore learns a unified correction operator over a family of BHE configurations in linearly heterogeneous media.

\paragraph{Distance indicator augmentation.}
We note that explicitly informing the neural networks the distance of sampled data points and collocation points from the location of the borehole benefits the approximation of near-source distribution. Thus, our model includes the horizontal distance to the reference borehole $\sqrt{x^2+y^2}$ as an input feature of the network. 

\paragraph{Selection of training points.}
To achieve a higher accuracy of the universal corrector approximation, we use a small number of supervised reference solutions as sparse anchors to calibrate the amplitude and early-time correction. The data supervision is integrated with physics-informed constraints which minimizes a combined residual loss over a set of sampled training points in space-time. To address the strong spatial and temporal scale imbalance exhibited in the correction field, both the supervised data points and the physics collocation points are selected using a source-adaptive and time-adaptive strategy.

Following the reformulation of the correction equation in (\ref{showing_coordinates_relative}) in terms of horizontal coordinates relative to the BHE location, the spatial sampling domain used for training is transformed accordingly into a relative coordinate system $\Omega_s^{'}$ for ILS-based model and $\Omega_s^{'}=\Omega_{xy}^{'}\times\left[z_{\min},z_{\max}\right]$ for FLS/MFLS-based model. To evaluate the PDE residuals, the horizontal coordinates of each collocation point are sampled adaptively according to its distance from the borehole. We aim to accurately resolve the steep local gradients near the source by increasing the distribution density of residual points in its vicinity. Here the horizontal plane is divided into near-source, intermediate and far-field regions by
\[
r_{\mathrm{core}}<r\le r_1,\qquad
r_1<r\le r_2,\qquad
r>r_2,
\]
and fixed portions of collocation points $\alpha_0$, $\alpha_1$, $\alpha_2$ are assigned to these regions. Points inside the immediate line source core are excluded from the PDE residual evaluation to avoid the singularity. One can alternatively adopt a polar sampling scheme \cite{Kara2025DeepONetFEM} to control the sampling density by adaptively partitioning the radial and angular directions.

For the FLS-based and MFLS-based models, the vertical coordinate is sampled independently with additional enrichment near the two endpoints of the finite line source $z=D$ and $z=D+L$. A mixture sampler with fixed portions $\beta_0$, $\beta_1$ and $\beta_2$ allocated to the endpoints, interior of line source and uniformly sampled full vertical domain.

To balance early transient behavior and long-term evolution, the full simulation horizon is divided into several time intervals $\{I_k\}_{k=1}^{N_t}$, and each collocation time is drawn uniformly from these intervals with prescribed probabilities
\[
t\sim \sum_{k=1}^{N_t} p_k\,\mathcal{U}(I_k),
\qquad
\sum_{k=1}^{N_t}p_k=1 .
\]

In addition to interior residual points, we generate collocation points to impose initial condition for three analytical-based models. 

The supervised data points are sampled following a similar procedure, except that only the BHE coordinates are excluded in the near-source region. Logarithmic sampling is employed in time variable to provide a denser distribution at early times. 

Therefore, the training points set for ILS-based model is 
\begin{align}
    \mathcal{S}_{\mathrm{st}}^{\mathrm{ILS}}
    =\mathcal{S}_{\mathrm{pde}}^{\mathrm{ILS}}\bigcup\mathcal{S}_{\mathrm{ic}}^{\mathrm{ILS}}\bigcup\mathcal{S}_{\mathrm{data}}^{\mathrm{ILS}},\notag
\end{align}
where
\begin{align}
    \mathcal{S}_{\mathrm{pde}}^{\mathrm{ILS}}=\bigl\{\left(x_i,y_i,t_i\right)|\left(x_i,y_i\right)\sim p_r\left(x,y\right),t_i\sim \sum_{k=1}^{N_t} p_k\,\mathcal{U}(I_k)\bigl\}_{i=1}^{N_{\mathrm{pde}}},\notag
\end{align}
with 
\begin{align}
p_r\left(x,y\right)=\alpha_0\mathcal{U}\!\left(\Omega^{'}_{s,r_\mathrm{core}<r\le r_1}\right)+\alpha_1\mathcal{U}\!\left(\Omega^{'}_{s,r_1<r\le r_2}\right)+\alpha_2\mathcal{U}\!\left(\Omega^{'}_{s,r>r_2}\right),\notag
\end{align}
\begin{align}
\mathcal{S}_{\mathrm{ic}}^{\mathrm{ILS}}=\Bigl\{\left(x_i,y_i,0\right)|\left(x_i,y_i\right)\sim\mathcal{U}
\!\left(
\Omega^{'}_{s,r>r_{\mathrm{core}}}
\right)\Bigr\}_{i=1}^{N_{\mathrm{ic}}},\notag
\end{align}
\begin{align}
    \mathcal{S}_{\mathrm{data}}^{\mathrm{ILS}}=\bigl\{\left(x_i,y_i,t_i\right)|\left(x_i,y_i\right)\sim p'_r\left(x,y\right),t_i\sim p_{\mathrm{log}}\left(t\right)\bigl\}_{i=1}^{N_{\mathrm{data}}},\notag
\end{align}
with
\begin{align}
    p'_r\left(x,y\right)=\alpha_0\mathcal{U}\!\left(\Omega^{'}_{s,0\neq r\le r_1}\right)+\alpha_1\mathcal{U}\!\left(\Omega^{'}_{s,r_1<r\le r_2}\right)+\alpha_2\mathcal{U}\!\left(\Omega^{'}_{s,r>r_2}\right),\notag
\end{align}
and
\begin{align}
p_{\log}(t)
=
\frac{1}{t\ln(T/t_\mathrm{core})},
\qquad
t\in[t_{\mathrm{core}},T].\notag
\end{align}

And the training point set for FLS-based and MFLS-based models is formulated as
\begin{align}
    \mathcal{S}_{\mathrm{st}}^{\mathrm{FLS}/\mathrm{MFLS}}
    =\mathcal{S}_{\mathrm{pde}}^{\mathrm{FLS}/\mathrm{MFLS}}\bigcup\mathcal{S}_{\mathrm{ic}}^{\mathrm{FLS}/\mathrm{MFLS}}\bigcup\mathcal{S}_{\mathrm{data}}^{\mathrm{FLS}/\mathrm{MFLS}},\notag
\end{align}
where
\begin{align}
    \mathcal{S}_{\mathrm{pde}}^{\mathrm{FLS}/\mathrm{MFLS}}=\bigl\{\left(x_i,y_i,z_i,t_i\right)|\left(x_i,y_i\right)\sim p_r\left(x,y\right),z_i\sim p_z\left(z\right),t_i\sim \sum_{k=1}^{N_t} p_k\,\mathcal{U}(I_k)\bigl\}_{i=1}^{N_{\mathrm{pde}}},\notag
\end{align}
with 
\begin{align}
    p_z\left(z\right)=\beta_0\mathcal{U}\!\left([D-r_z,D+r_z]\cup[D+L-r_z,D+L+r_z]\right)+\beta_1\mathcal{U}\!\left([D,D+L]\right)
    +\beta_2\mathcal{U}\!\left([z_{\min},z_{\max}]\right),\notag
\end{align}

\begin{align}
\mathcal{S}_{\mathrm{ic}}^{\mathrm{FLS}/\mathrm{MFLS}}=\bigl\{\left(x_i,y_i,z_i,0\right)|
\left(x_i,y_i\right)\sim p_r\left(x,y\right),
z_i\sim p_z\left(z\right)
\Bigr\}_{i=1}^{N_{\mathrm{ic}}},\notag
\end{align}

\begin{align}
    \mathcal{S}_{\mathrm{data}}^{\mathrm{FLS}/\mathrm{MFLS}}=\bigl\{\left(x_i,y_i,z_i,t_i\right)|\left(x_i,y_i\right)\sim p'_r\left(x,y\right),z_i\sim p_z\left(z\right),t_i\sim p_{\log}\left(t\right)\bigl\}_{i=1}^{N_{\mathrm{data}}}.\notag
\end{align}

We denote the linear parametrization for the local conductivity as a parameter vector $\boldsymbol{\mu}=\left(\lambda_B,\lambda_x,\lambda_y\right)$, and sample the training vectors in parametric space $\mathcal{P}=\bigl\{\boldsymbol{\mu}:\lambda_B\in\left[\lambda_B^{\min},\lambda_B^{\max}\right],\lambda_x\in\left[\lambda_x^{\min},\lambda_x^{\max}\right],\lambda_y\in\left[\lambda_y^{\min},\lambda_y^{\max}\right]\bigl\}$ to evaluate the solutions corresponding to sampled thermal conditions on training points.
For the supervised data points, we generate high-fidelity reference solutions using the finite difference method for randomly selected conductivity parameters $\boldsymbol{\mu}^{(j)}\in\mathcal{P}$. The reference solutions are derived using a conservative finite difference scheme on a Cartesian  grid with implicit time integration, where the singular source is discretized over the computational grid nodes or cells while preserving the prescribed heat input. For each sampled parameter realization, the supervised label is computed as 
$\widetilde{T}_{\mathrm{data},c}^m\left(\mathbf{x},t;\boldsymbol{\mu}^{\left(j\right)}\right)=T_{h}\left(\mathbf{x},t;\boldsymbol{\mu}^{\left(j\right)}\right)-T_u-\widetilde{T}^m\left(\mathbf{x},t;\lambda_0\right)$, $m \in \{\mathrm{ILS},\mathrm{FLS},\mathrm{MFLS}\}$, where $T_{h}\left(\mathbf{x},t;\boldsymbol{\mu}^{\left(j\right)}\right)$ is the absolute temperature obtained from the FDM solver with the borehole at the horizontal origin. For the MFLS model, the advection velocity \(\mathbf v=(v_x,v_y,0)\) is also specified in the analytical baseline.

Thus, the parametric neural network is trained to approximate a family of correction fields over the parameter space $\mathcal{P}$. Combining the training points and their distance indicator augmentation, we define the loss function for ILS-based model as
\begin{equation}
\mathcal{L}^{\mathrm{ILS}}
=
w_{\mathrm{data}}^{\mathrm{ILS}}\mathcal{L}_{\mathrm{data}}^{\mathrm{ILS}}
+
w_{\mathrm{pde}}^{\mathrm{ILS}}\mathcal{L}_{\mathrm{pde}}^{\mathrm{ILS}}
+
w_{\mathrm{ic}}^{\mathrm{ILS}}\mathcal{L}_{\mathrm{ic}}^{\mathrm{ILS}},\notag
\end{equation}
where
\begin{equation}
\mathcal{L}_{\mathrm{data}}^{\mathrm{ILS}}
=
\frac{1}{N_{p,1}N_{\mathrm{data}}}
\sum_{j=1}^{N_{p,1}}
\sum_{i=1}^{N_{\mathrm{data}}}
\left|
\frac{
\widetilde{T}^{\mathrm{ILS}}_{\theta,c}(x_i,y_i,t_i,r_i;\boldsymbol{\mu}^{\left(j\right)})
-
\widetilde{T}^{\mathrm{ILS}}_{\mathrm{data},c}\left(x_i,y_i,t_i;\boldsymbol{\mu}^{\left(j\right)}\right)}{C_{\mathrm{scale}}}\right|^2,\notag
\end{equation}
\begin{equation}
\mathcal{L}_{\mathrm{pde}}^{\mathrm{ILS}}
=
\frac{1}{N_{p,2}N_{\mathrm{pde}}}
\sum_{j=1}^{N_{p,2}}
\sum_{i=1}^{N_{\mathrm{pde}}}
\left|\frac{\left[\rho_sC_s\frac{\partial \widetilde{T}^{\mathrm{ILS}}_{\theta,c}}{\partial t}-\nabla\cdot\left(\lambda^{\left(j\right)}\nabla\widetilde{T}^{\mathrm{ILS}}_{\theta,c}\right)-\nabla\cdot\left(\left(\lambda^{\left(j\right)}-\lambda_0\right)\nabla\widetilde{T}^{\mathrm{ILS}}\right)\right]\left(x_i,y_i,t_i,r_i;\boldsymbol{\mu}^{\left(j\right)}\right)}{R_{\mathrm{scale}}}\right|^2 ,\notag
\end{equation}
and
\begin{equation}
\mathcal{L}_{\mathrm{ic}}^{\mathrm{ILS}}
=
\frac{1}{N_{p,3}N_{\mathrm{ic}}}
\sum_{j=1}^{N_{p,3}}
\sum_{i=1}^{N_{\mathrm{ic}}}
\left|
\frac{
\widetilde{T}_{\theta,c}^{\mathrm{ILS}}\left(x_i,y_i,0,r_i;\boldsymbol{\mu}^{\left(j\right)}\right)
}{C_{\mathrm{scale}}}\right|^2 ,\notag
\end{equation}
with predefined normalization scales $C_{\mathrm{scale}}$, $R_{\mathrm{scale}}$, $\lambda^{\left(j\right)}=\lambda_B^{\left(j\right)}+\lambda_x^{\left(j\right)}x+\lambda_y^{\left(j\right)}y$, $ r_i=\sqrt{x_i^2+y_i^2}$ and the numbers of parameter realizations $N_{p,1}$ for data loss and $N_{p,2}$, $N_{p,3}$ for the PDE and initial condition losses, respectively. 

The loss function for FLS-based model is formulated as
\begin{equation}
\mathcal{L}^{\mathrm{FLS}}
=
w_{\mathrm{data}}^{\mathrm{FLS}}\mathcal{L}_{\mathrm{data}}^{\mathrm{FLS}}
+
w_{\mathrm{pde}}^{\mathrm{FLS}}\mathcal{L}_{\mathrm{pde}}^{\mathrm{FLS}}
+
w_{\mathrm{ic}}^{\mathrm{FLS}}\mathcal{L}_{\mathrm{ic}}^{\mathrm{FLS}},\notag
\end{equation}
where
\begin{equation}
\mathcal{L}_{\mathrm{data}}^{\mathrm{FLS}}
=
\frac{1}{N_{p,1}N_{\mathrm{data}}}
\sum_{j=1}^{N_{p,1}}
\sum_{i=1}^{N_{\mathrm{data}}}
\left|
\frac{
\widetilde{T}^{\mathrm{FLS}}_{\theta,c}(x_i,y_i,z_i,t_i,r_i;\boldsymbol{\mu}^{\left(j\right)})
-
\widetilde{T}^{\mathrm{FLS}}_{\mathrm{data},c}\left(x_i,y_i,z_i,t_i;\boldsymbol{\mu}^{\left(j\right)}\right)}{C_{\mathrm{scale}}}\right|^2,\notag
\end{equation}
\begin{align}
\mathcal{L}_{\mathrm{pde}}^{\mathrm{FLS}}
=&
\frac{1}{N_{p,2}N_{\mathrm{pde}}}
\sum_{j=1}^{N_{p,2}}
\sum_{i=1}^{N_{\mathrm{pde}}}\notag\\
&\left|\frac{\left[\rho_sC_s\frac{\partial \widetilde{T}^{\mathrm{FLS}}_{\theta,c}}{\partial t}-\nabla\cdot\left(\lambda^{\left(j\right)}\nabla \widetilde{T}^{\mathrm{FLS}}_{\theta,c}\right)-\nabla\cdot\left(\left(\lambda^{\left(j\right)}-\lambda_0\right)\nabla\widetilde{T}^{\mathrm{FLS}}\right)\right]\left(x_i,y_i,z_i,t_i,r_i;\boldsymbol{\mu}^{\left(j\right)}\right)}{R_{\mathrm{scale}}}\right|^2 ,\notag
\end{align}
and
\begin{equation}
\mathcal{L}_{\mathrm{ic}}^{\mathrm{FLS}}
=
\frac{1}{N_{p,3}N_{\mathrm{ic}}}
\sum_{j=1}^{N_{p,3}}
\sum_{i=1}^{N_{\mathrm{ic}}}
\left|
\frac{
\widetilde{T}_{\theta,c}^{\mathrm{FLS}}\left(x_i,y_i,z_i,0,r_i;\boldsymbol{\mu}^{\left(j\right)}\right)
}{C_{\mathrm{scale}}}\right|^2 .\notag
\end{equation}
Similarly for MFLS-based model, we have
\begin{equation}
\mathcal{L}^{\mathrm{MFLS}}
=
w_{\mathrm{data}}^{\mathrm{MFLS}}\mathcal{L}_{\mathrm{data}}^{\mathrm{MFLS}}
+
w_{\mathrm{pde}}^{\mathrm{MFLS}}\mathcal{L}_{\mathrm{pde}}^{\mathrm{MFLS}}
+
w_{\mathrm{ic}}^{\mathrm{MFLS}}\mathcal{L}_{\mathrm{ic}}^{\mathrm{MFLS}},\notag
\end{equation}
where $\mathcal{L}_{\mathrm{data}}^{\mathrm{MFLS}}$ follows the same definition as $\mathcal{L}_{\mathrm{data}}^{\mathrm{FLS}}$, 
\begin{align}
\mathcal{L}_{\mathrm{pde}}^{\mathrm{MFLS}}
=
\frac{1}{N_{p,2}N_{\mathrm{pde}}}
\sum_{j=1}^{N_{p,2}}
\sum_{i=1}^{N_{\mathrm{pde}}}
\left|\frac{\mathcal{R}\left(x_i,y_i,z_i,t_i,r_i;\boldsymbol{\mu}^{\left(j\right)}\right)}{R_{\mathrm{scale}}}\right|^2 ,\notag
\end{align}
with
\begin{align}
    \mathcal{R}=\rho C \frac{\partial \widetilde{T}^{\mathrm{MFLS}}_{\theta,c}}{\partial t}
+\rho_f C_f\mathrm{\mathbf{v}}\cdot\nabla \widetilde{T}^{\mathrm{MFLS}}_{\theta,c}-\nabla\cdot\left(\lambda^{\left(j\right)}\nabla \widetilde{T}^{\mathrm{MFLS}}_{\theta,c}\right)-\nabla\cdot\left(\left(\lambda^{\left(j\right)}-\lambda_0\right)\nabla\widetilde{T}^{\mathrm{MFLS}}\right),\notag
\end{align}
and
\begin{equation}
\mathcal{L}_{\mathrm{ic}}^{\mathrm{MFLS}}
=
\frac{1}{N_{p,3}N_{\mathrm{ic}}}
\sum_{j=1}^{N_{p,3}}
\sum_{i=1}^{N_{\mathrm{ic}}}
\left|
\frac{
\widetilde{T}_{\theta,c}^{\mathrm{MFLS}}\left(x_i,y_i,z_i,0,r_i;\boldsymbol{\mu}^{\left(j\right)}\right)
}{C_{\mathrm{scale}}}\right|^2.\notag
\end{equation}

\subsection{Spatial and Temporal Superposition}

As discussed above, the trained parametric corrector provides the thermal response for a single BHE with unit line load in a computational domain with linearly changing conductivity field. For an arbitrary borehole
located at \(\left(x_{\mathrm{BHE}}^k,y_{\mathrm{BHE}}^k\right)\), the unit load hybrid response is written as
\[
T_{\theta^*,k}^m\left(\mathbf{x},t;\boldsymbol{\mu}_k\right)
=T_u+
\widetilde{T}^m(\mathbf{x},t;\lambda_0)
+
\widetilde{T}^m_{\theta^*,c}(x-x_{\mathrm{BHE}}^k,y-y_{\mathrm{BHE}}^k,z,t,r;\boldsymbol{\mu}_k),
\]
where $\boldsymbol{\mu}_k=\left(\lambda\left(x_{\mathrm{BHE}}^k,y_{\mathrm{BHE}}^k\right),\lambda_x\left(x_{\mathrm{BHE}}^k,y_{\mathrm{BHE}}^k\right),\lambda_y\left(x_{\mathrm{BHE}}^k,y_{\mathrm{BHE}}^k\right)\right)$, $\theta^*$ are the updated trainable parameters of the network and $T_{k}^m\left(\mathbf{x},t;\boldsymbol{\mu}_k\right)$ denotes the predicted soil temperature with regard to analytical model $m\in\{\mathrm{ILS},\mathrm{FLS},\mathrm{MFLS}\}$. For the two-dimensional ILS model, the \(z\)-dependent terms are omitted. 

With the assumption that the soil thermal properties do not depend on the temperature distribution, the governing heat diffusion equation, advection-diffusion equation and the analytical solutions are linear with respect to the source load for a prescribed conductivity field. Due to the additivity of the thermal energy, the response to multiple BHEs and time-varying loads can be reconstructed by spatial and temporal superposition \cite{dePaly2012Optimization}. Let the load history be discretized into $N_t$ piecewise-constant time intervals,
with
\[
t_l=l\Delta t,\qquad l=0,\ldots,N_t,
\]
and let \(q_{k,l}\) denote the constant load of the \(k\)-th borehole during \([t_{l},t_{l+1})\), $l=0,1,\cdots,N_t-1$, with $q_{k,-1}=0$. The predicted temperature response at time step \(t_{N_t-1}<t\le t_{N_t}\) concerning $K$ BHEs in the computation domain with a certain conductivity $\lambda\left(x,y\right)$ is calculated as
\begin{align}    &T^m_{\theta^*}\left(\mathbf{x},t;\lambda\left(x,y\right)\right)
    =T_{u}\notag+\sum_{k=1}^K\sum_{l=0}^{N_t-1}\left(\widetilde{T}^m_{k,l}\left(\mathbf{x},t;\lambda_0\right)+ \left(q_{k,l}-q_{k,l-1}\right)\widetilde{T}^m_{\theta^*,c}\left(x-x_{\mathrm{BHE}}^k,y-y_{\mathrm{BHE}}^k,z,t-t_l,r;\boldsymbol{\mu}_k\right)\right).\notag
\end{align}
Here, $\widetilde{T}^m_{k,l}$ denotes the single step analytical approximation of model $m$ concerning the load $q_{k,l}-q_{k,l-1}$ and the thermal conductivity $\lambda_0$ in (\ref{eq:ils_piecewise_load}), (\ref{eq:fls_piecewise_load}) and (\ref{eq:mfls_piecewise_load}). The superposition principle allows the trained parametric universal corrector $\widetilde{T}^m_{\theta^*,c}$ to be used for multi-BHE field simulations without retraining the neural network for various borehole locations and thermal heterogeneities. 

\begin{figure}[b!]
    \centering
    \includegraphics[width=0.95\textwidth]{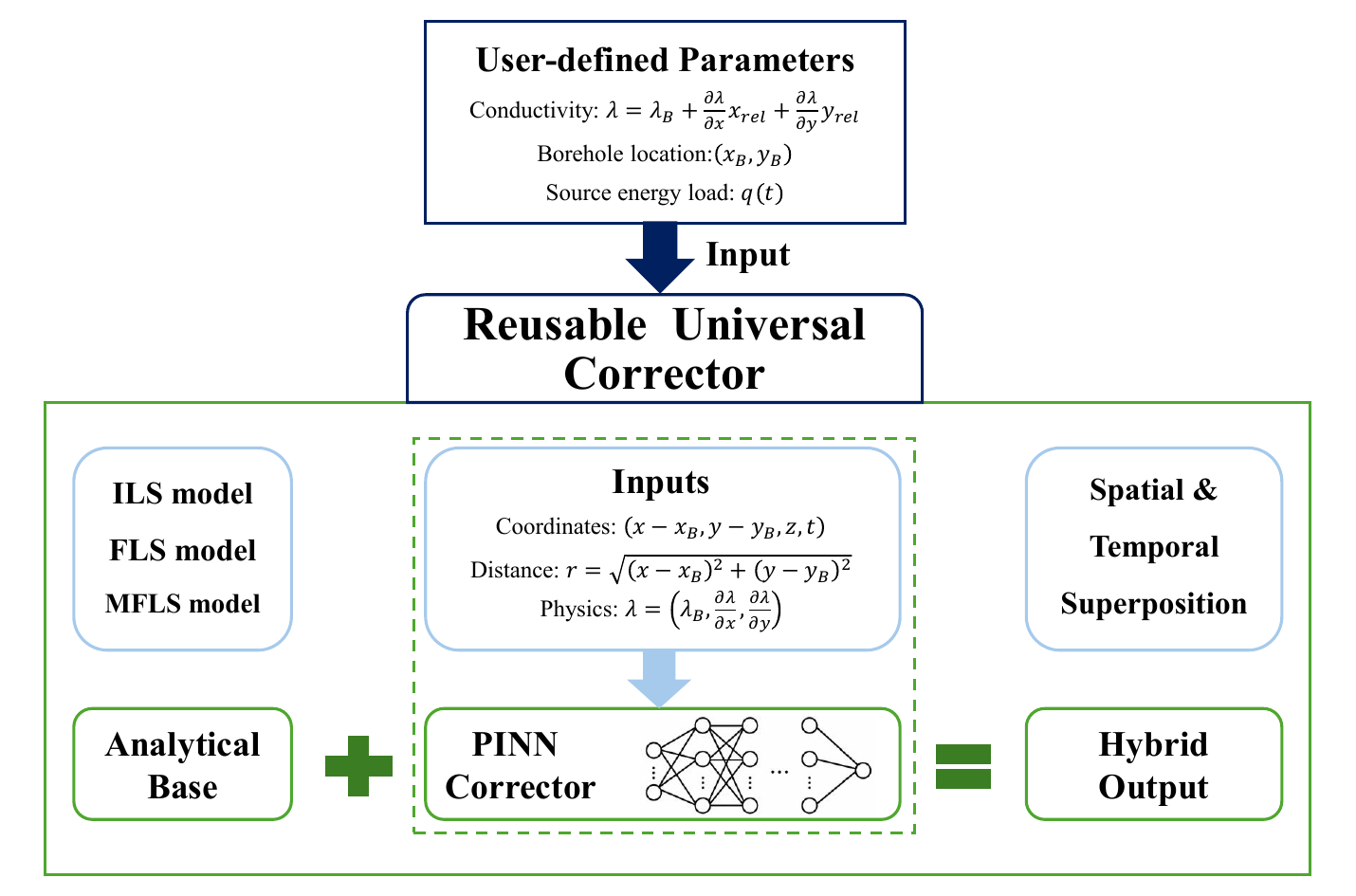}
    \caption{Schematic framework of the reusable universal PINN corrector. The framework outlines the proposed hybrid analytical-PINN approach, which utilizes line source analytical base to learn the relatively regular correction by physics-informed fully connected neural network.}
    \label{fig:schematic_flowchart}
\end{figure}

The proposed hybrid analytical-PINN model framework is schematically presented in Figure \ref{fig:schematic_flowchart}. Inspired by the splitting concept \cite{hu2023singularitysplitting}, our model extracts the line source singularity and learns solution corresponding to specific parametrized thermal conductivity. By reformulating the correction equation with respect to BHE-centered relative coordinates, using unit source load during training, and parametrizing the conductivity field into a gradient formulation, the corrector network is made universal and reusable. Training the model under a unit load allows the predicted correction to be rescaled and superimposed for arbitrary time-dependent source load profiles. The parametric PINN enables the corrector to account for heterogeneous soil conditions within the prescribed range. The relative coordinate representation facilitates application to various BHE arrangements. Combining these, a single trained corrector can be reused to evaluate a broad range of BHE design and operating conditions without retraining. The trained model can calculate the correction at any spatial location and time step within the trained ranges with less amount of computation time than the numerical models. In summary, the hybrid model provides a computationally efficient framework for repeated evaluation of BHE systems under varying heterogeneous soil conditions, borehole configurations and load profiles.

\section{Case Study}\label{case_study}
In this section, to illustrate the effectiveness of the proposed parametric hybrid-PINN corrector method, we present several numerical results in two and three spatial dimensions for BHE-induced soil thermal problem. After the training process is finished, we examine the accuracy of predicted hybrid solution by computing the relative $L^2$ and $L^{\infty}$ errors between the learned solution and reference solution obtained from the conservative finite difference solver described in Section \ref{parametric_pinn_method}. To be specific, we use the visualization grid points as the test points to evaluate the errors of corrections based on a certain analytical model $m$ at a selected time step $t$ by
\begin{align}
\mathcal{E}_{\infty,\mathrm{hyb},m}^{t}=\frac{\|T^m_{\theta^*}-T_h\|_{\infty}}{\|T_h\|_{\infty}},\qquad
\mathcal{E}_{2,\mathrm{hyb},m}^{t}=\frac{\|T^m_{\theta^*}-T_h\|_{2}}{\|T_h\|_{2}},\notag
\end{align}
respectively, where 
\begin{equation}
    \|T_h\|_{\infty}=\max\limits_{1\le i\le N_{\mathrm{test}}}|T_h\left(\mathbf{x}_i,t\right)|,\qquad
    \|T_h\|_{2}=\sqrt{\frac{1}{N_{\mathrm{test}}}\sum_{i=1}^{N_{\mathrm{test}}}T_h^2\left(\mathbf{x}_i,t\right)}.\notag
\end{equation}
For comparison, we consider two analytical baselines for multi-BHE systems. In addition to the conventional analytical solution using a single bulk conductivity, we introduce a local conductivity approximation in which the response of each borehole is evaluated using the thermal conductivity at its own location before spatial superposition.
\begin{align}
    \widetilde{T}^m\left(\mathbf{x},t\right)=\sum_{k=1}^K\sum_{l=0}^{N_t-1}\widetilde{T}^m_{k,l}\left(\mathbf{x},t;\lambda_0\right),\qquad
    \widetilde{T}^m_{lo}\left(\mathbf{x},t\right)=\sum_{k=1}^K\sum_{l=0}^{N_t-1}\widetilde{T}^m_{k,l,lo}\left(\mathbf{x},t;\lambda\left(x_{\mathrm{BHE}}^k,y_{\mathrm{BHE}}^k\right)\right),\notag
\end{align}
where $\widetilde{T}^m_{k,l}$ represents the single step response of model $m$ corresponding to the load increment $q_{k,l}-q_{k,l-1}$ and the bulk conductivity $\lambda_0$. $\widetilde{T}^m_{k,l,lo}$ denotes the response of the $k$-th borehole using its local conductivity $\lambda\left(x_{\mathrm{BHE}}^k,y_{\mathrm{BHE}}^k\right)$. Hereafter, the multi-BHE solutions obtained using the bulk conductivity and the local conductivities are referred to as the global and local analytical solution, respectively. The corresponding errors are computed by
\begin{align}
\mathcal{E}_{\infty,\mathrm{ana},m}^{t}=\frac{\|T_u+\widetilde{T}^m-T_h\|_{\infty}}{\|T_h\|_{\infty}},\qquad
\mathcal{E}_{2,\mathrm{ana},m}^{t}=\frac{\|T_u+\widetilde{T}^m-T_h\|_{2}}{\|T_h\|_{2}}.\notag
\end{align}
Substituting $\widetilde{T}^m$ by $\widetilde{T}_{lo}^m$ gives the errors for locally estimated analytical solutions. For each selected time step, the FDM reference solution is interpolated onto the same evaluation grid as the analytical and hybrid predictions. In general, with 
\(\Delta x=\Delta y=\Delta z=1\,\mathrm{m}\), we set \(N_{\mathrm{test}}=81\times121=9801\) or $N_{\mathrm{test}}=114\times121=13794$ points for vertical slice (a single two dimensional vertical cross-sectional plane within the computational domain) visualization depending on the scenario considered, and $N_{\mathrm{test}}=81\times81=6561$ for each horizontal slice. 

Throughout all numerical experiments for all three investigated base models, the neural corrector is implemented as a fully connected network with five hidden layers, 128 neurons per layer and \(\tanh\) activations. The linear output layer returns the correction field \(\widetilde{T}^{\mathrm{ILS}}_{\theta^*,c}\), $\widetilde{T}^{\mathrm{FLS}}_{\theta^*,c}$ and  $\widetilde{T}^{\mathrm{MFLS}}_{\theta^*,c}$. The network model is trained using AdamW optimizer with a learning rate of \(10^{-4}\), weight decay \(10^{-5}\) and gradient clipping with threshold \(0.5\). To train all three analytical solution based models, 32 numerical simulations with different conductivities are used to generate the correction data. For ILS-based example, 4096 supervised points are sampled from the corresponding FDM solution. In addition, 4096 collocation points are sampled at each iteration to evaluate the PDE residual, and 1024 points are used for the initial condition loss. For FLS and MFLS based cases, the numbers of training points for data, PDE and initial condition loss terms are set to 8192, 8192 and 2048, respectively. 

\begin{figure}[bhtp]
    \centering
    \begin{subfigure}[t]{0.38\textwidth}
        \centering
        \includegraphics[width=\textwidth]{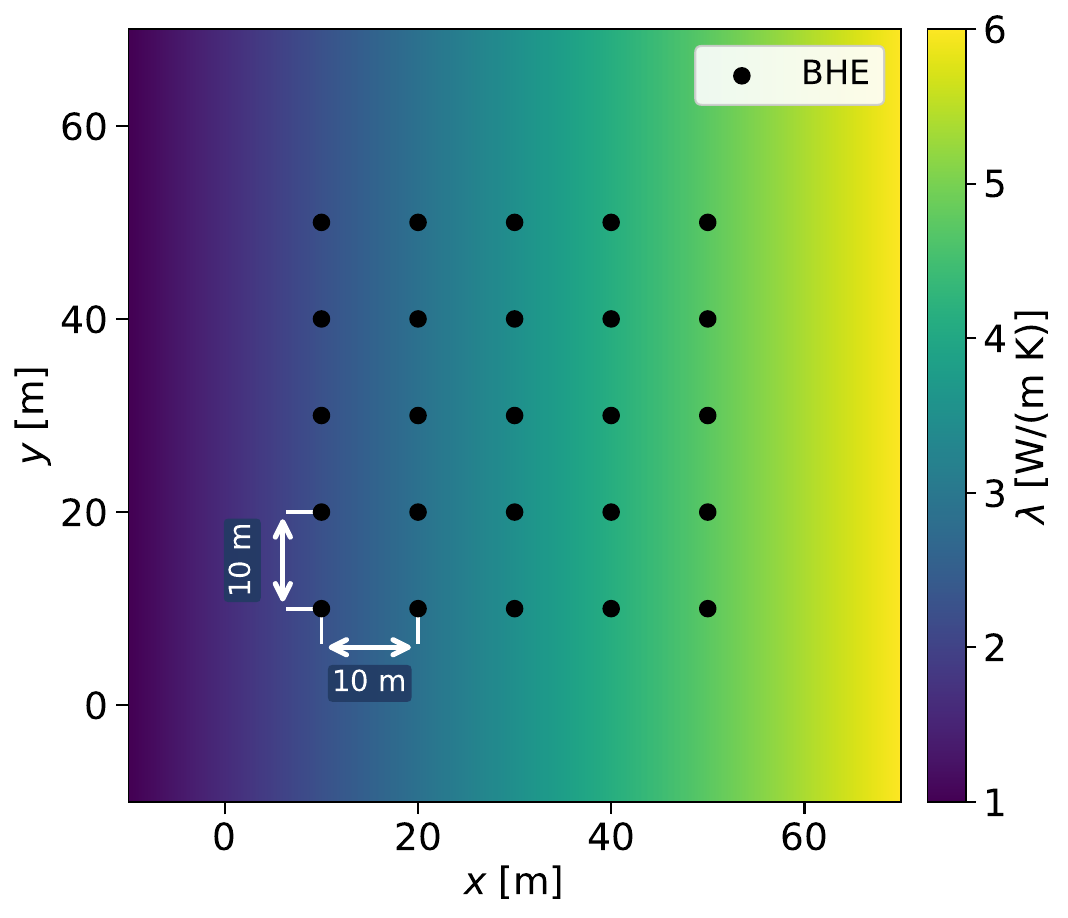}
        \caption{Scenario 1}
        \label{fig:kx}
    \end{subfigure}
    \hspace{5mm}
    \begin{subfigure}[t]{0.38\textwidth}
        \centering
        \includegraphics[width=\textwidth]{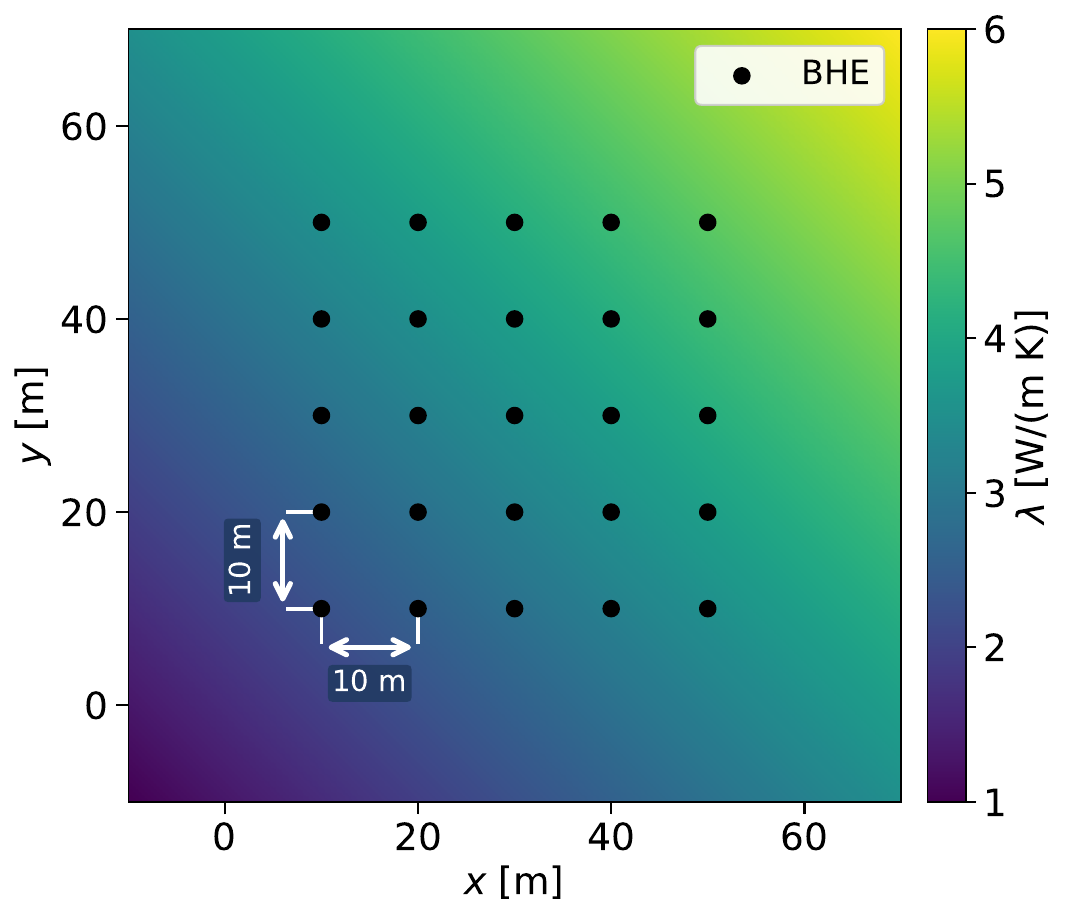}
        \caption{Scenario 2}
        \label{fig:kxy}
    \end{subfigure}
    \caption{Thermal conductivity fields considered for testing ILS, FLS and MFLS based models.}
    \label{fig:conductivity}
\end{figure}

In the following examples, we consider two thermal conductivity scenarios: one varying linearly along the $x$ direction and the other along both the $x$ and $y$ directions, as shown in Fig. \ref{fig:conductivity}. 
The remaining parameters featuring the soil and fluid for the ILS/FLS-based and MFLS-based models are listed in Tables \ref{tab:ils_fls_parameters} and \ref{tab:mfls_parameters}. 
\begin{table}[htbp]
\centering
\footnotesize
\caption{Physical property settings for demonstration examples used for training and inference procedure of parametric hybrid-PINN corrector method based on ILS and FLS models.}
\label{tab:ils_fls_parameters}

\begingroup
\renewcommand{\arraystretch}{1.20}
\setlength{\tabcolsep}{7pt}
\setlength{\aboverulesep}{0.6ex}
\setlength{\belowrulesep}{0.6ex}

\begin{tabular}{lll}
\toprule
Parameter & Value & Unit \\
\midrule
Solid density $\rho_s$
& $2650$
& $\mathrm{kg\,m^{-3}}$ \\

Solid specific heat capacity $C_s$
& $800$
& $\mathrm{J\,kg^{-1}\,K^{-1}}$ \\

Solid volumetric heat capacity $\rho_s C_s$
& $2.12\times10^{6}$
& $\mathrm{J\,m^{-3}\,K^{-1}}$ \\
\bottomrule
\end{tabular}
\endgroup
\end{table}

\begin{table}[htbp]
\centering
\footnotesize
\caption{Physical property settings for demonstration examples used for training and inference procedure of parametric hybrid-PINN corrector method based on MFLS model.}
\label{tab:mfls_parameters}

\begingroup
\renewcommand{\arraystretch}{1.20}
\setlength{\tabcolsep}{7pt}
\setlength{\aboverulesep}{0.6ex}
\setlength{\belowrulesep}{0.6ex}

\begin{tabular}{lll}
\toprule
Parameter & Value & Unit \\
\midrule
Groundwater density $\rho_f$
& $998.2$
& $\mathrm{kg\,m^{-3}}$ \\

Groundwater specific heat capacity $C_f$
& $4182$
& $\mathrm{J\,kg^{-1}\,K^{-1}}$ \\

Groundwater volumetric heat capacity $\rho_f C_f$
& $4.17\times10^{6}$
& $\mathrm{J\,m^{-3}\,K^{-1}}$ \\

Solid density $\rho_s$
& $2650$
& $\mathrm{kg\,m^{-3}}$ \\

Solid specific heat capacity $C_s$
& $800$
& $\mathrm{J\,kg^{-1}\,K^{-1}}$ \\

Solid volumetric heat capacity $\rho_s C_s$
& $2.12\times10^{6}$
& $\mathrm{J\,m^{-3}\,K^{-1}}$ \\

Effective porosity $\epsilon$
& $0.14$
& -- \\
\bottomrule
\end{tabular}
\endgroup
\end{table}

\begin{figure}[b!]
    \centering
    \includegraphics[width=0.95\textwidth]{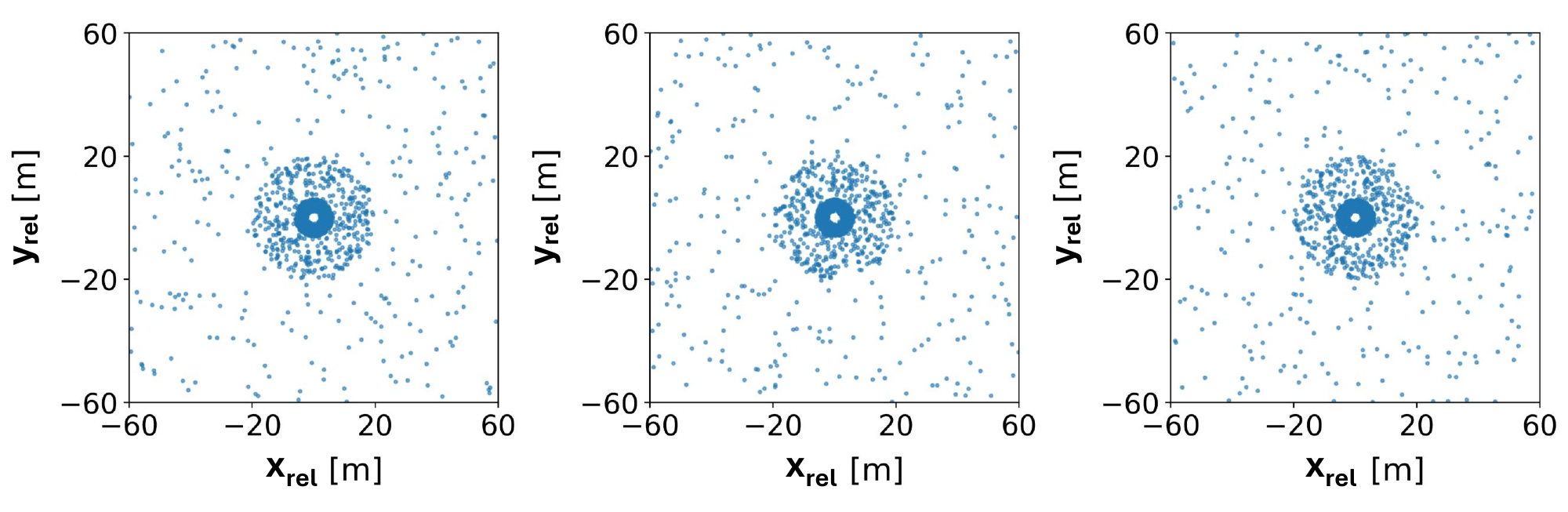}
    \caption{Visualization of PDE collocation points sampled for Example 1 in month 1, 60 and 360 (from left to right). }
    \label{fig:ils_sampling}
\end{figure}

\begin{figure}[tb!]
    \centering
    \begin{subfigure}[t]{0.4\textwidth}
        \centering
        \includegraphics[width=\textwidth]{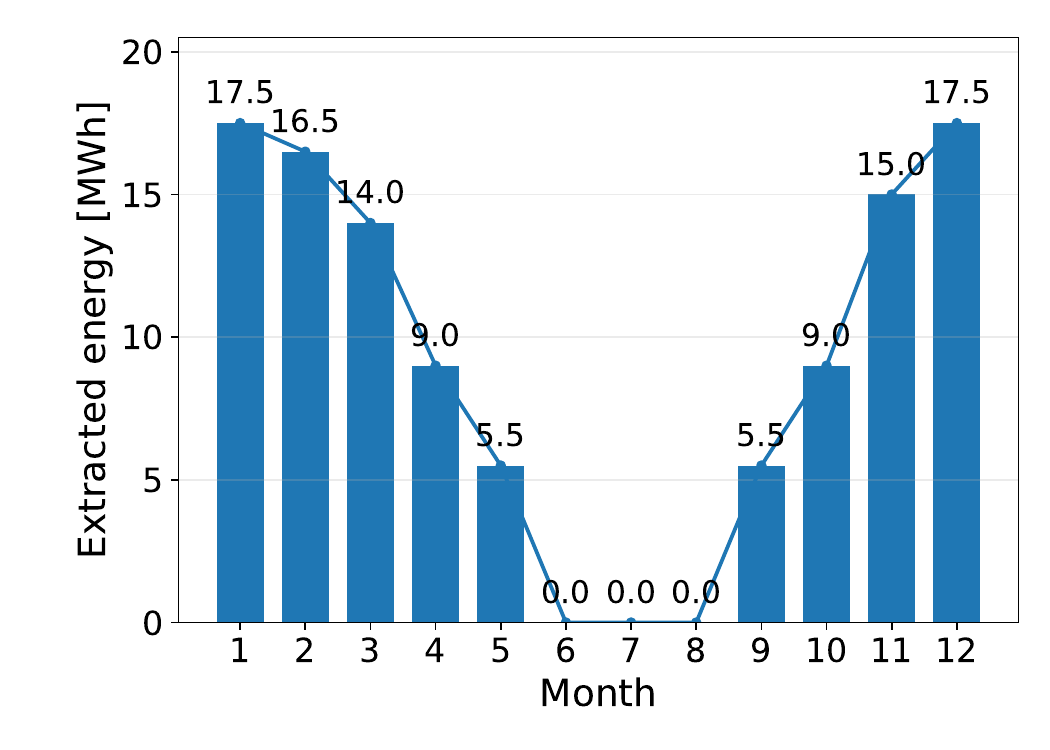}
        \caption{Extracted energy per month.}
    \end{subfigure}
    \hspace{5mm}
    \begin{subfigure}[t]{0.4\textwidth}
        \centering
        \includegraphics[width=\textwidth]{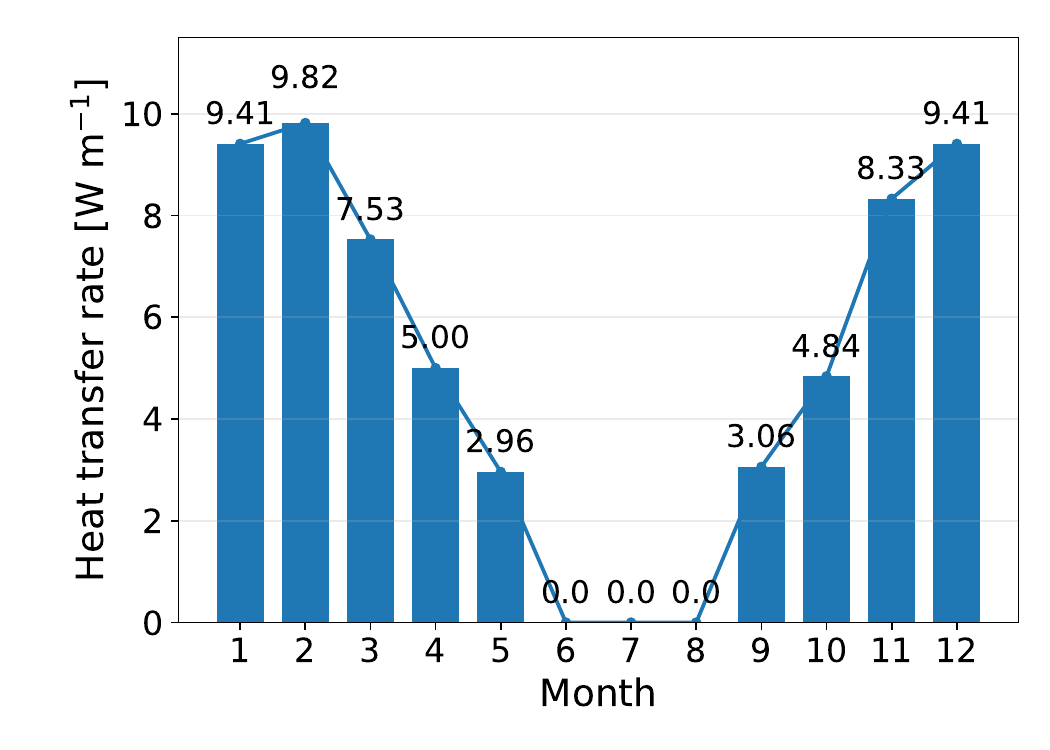}
        \caption{Heat transfer rate per month of the considered system consisting of 25 BHEs with identical lengths $L=100$m and heat transfer rates.}
    \end{subfigure}
    \caption{Annual source load profile repeatedly used in the simulations (based on \cite{dePaly2012Optimization}).}
    \label{fig:load profile}
\end{figure}

\paragraph{Example 1.(ILS-based Model)}
We choose a square domain $\Omega_s=\left[-10,70\right]\mathrm{m}\times\left[-10,70\right]\mathrm{m}$ and a simulation time span of $30$ years, and we consider a case where $25$ BHEs are arranged as a regular $5\times5$ array with a spacing $10$m. In terms of model training, we derive the bulk thermal conductivity $\lambda_0$ as the average of the conductivities at all locations of the boreholes. To test the proposed model on the two scenarios, we introduce an interpolation parameter $\gamma\sim\mathcal{U}(0,1)$ to be sampled. Let \(\lambda_{\min}^{\mathrm{train}}=\lambda_{\mathrm{min}}\), \(\lambda_{\max}^{\mathrm{train}}=\lambda_{\mathrm{max}}\) and the conductivity gradient parameters are defined as
\[
\lambda_x(\gamma)
=
(1-\gamma)
\frac{\lambda_{\max}^{\mathrm{train}}-\lambda_{\min}^{\mathrm{train}}}
{x_{\max}-x_{\min}},\qquad
\lambda_y(\gamma)
=
\gamma
\frac{\lambda_{\max}^{\mathrm{train}}-\lambda_{\min}^{\mathrm{train}}}
{y_{\max}-y_{\min}}.
\]
The selected conductivity parameter is then derived by combining
\begin{align}
    \left(\lambda_B\left(\gamma\right),\lambda_x\left(\gamma\right),\lambda_y\left(\gamma\right)\right),\notag
\end{align}
where
\begin{align}
    \lambda_B\left(\gamma\right)=\lambda(x_{\mathrm{BHE}},y_{\mathrm{BHE}};\gamma)
=
\lambda_{\min}^{\mathrm{train}}
+
\lambda_x(\gamma)(x_{\mathrm{BHE}}-x_{\min})
+
\lambda_y(\gamma)(y_{\mathrm{BHE}}-y_{\min}).\notag
\end{align}
The training samples are generated in the predefined computational domain under the sampled conductivity conditions $\boldsymbol{\mu}\left(\gamma\right)$ with the borehole fixed at the horizontal origin. We set $N_{p,1}=32$ and $N_{p,2}=N_{p,3}=1000$ to construct the data and physics loss terms. The spatial coordinates of physics collocation points and the interior data points are both chosen using a three-layer adaptive strategy around the BHE (see Section \ref{parametric_pinn_method}). We use
\begin{align}
    r_{\mathrm{core}}=2~\mathrm{m},\qquad
r_1=6~\mathrm{m},\qquad
r_2=20~\mathrm{m},\notag
\end{align}
with portions
\begin{align}
    \alpha_0=0.50,\qquad
\alpha_1=0.3,\qquad
\alpha_2=0.2 .\notag
\end{align}
The PDE collocation times are sampled from four stratified intervals \([1,24]\), \([24,72]\), \([72,180]\) and \([180,360]\) months, with equal probability \(0.25\) for each interval. Figure \ref{fig:ils_sampling} illustrates the sampled collocation points for the PDE residual. During the inference, we adopt the source load profile from \cite{dePaly2012Optimization} where the energy extraction distribution for 1 year is given and we apply the monthly adapted pattern as shown in Fig. \ref{fig:load profile} repeatedly over the 30-year period. A uniform heat rate per unit length is assumed along each active borehole
segment. The trained universal corrector is then evaluated for each BHE using its own local parameter vector \(\boldsymbol{\mu}_k\left(k=1,...,25\right)\) regarding two scenarios. The multi-BHE field response is obtained by spatial superposition over the \(25\) BHEs and temporal superposition over the piecewise-constant monthly load history.

The hybrid-PINN model is able to recover the heterogeneous temperature field with high accuracy for both conductivity scenarios. As summarized in Table \ref{tab:ils_x_15_open_relative_errors}, the hybrid solution achieves consistently higher accuracy than both the global and local analytical approximations, indicating that the learned correction captures not only the local temperature response around individual boreholes but also the large-scale bias induced by subsurface heterogeneity. Specifically, the relative \(L^2\) error of the hybrid solution remains below approximately $1.7\%$ for both scenarios, while the \(L^\infty\) error stays around $3\%
$. Moreover, whereas the errors of the analytical solutions increase noticeably with simulation time, the hybrid solution exhibits much smaller temporal variation, demonstrating a more stable long-term approximation.

\begin{figure}[!htbp]
    \centering
    \includegraphics[width=0.95\textwidth,
    trim=0 4.6cm 0 4.6cm,
    clip]{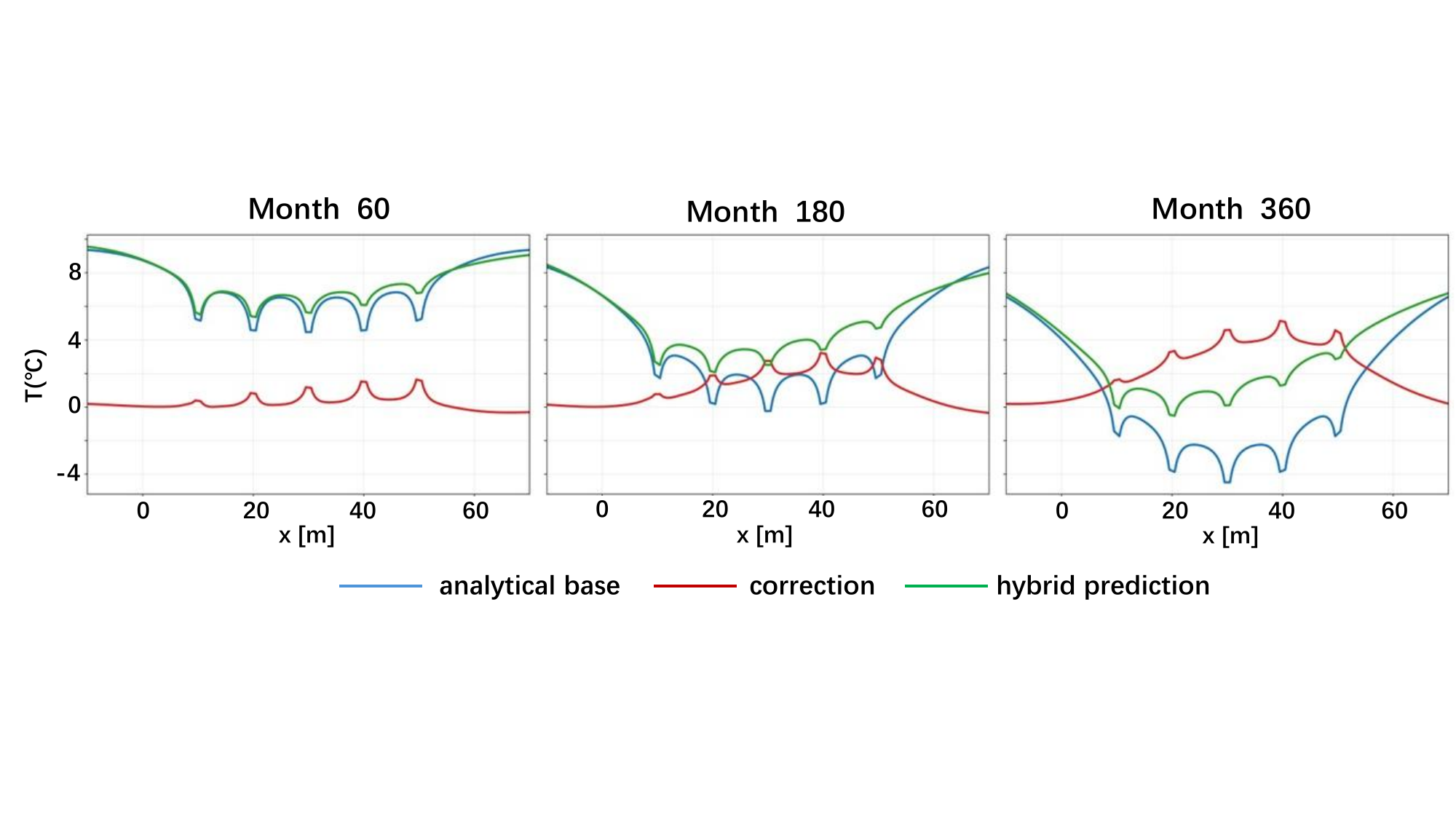}
    \caption{Visualization of the role that neural correction plays for temperature simulation in heterogeneous soil in Decembers of years 5, 15, and 30 (from left to right). Each panel depicts: the temperature distribution modeled by ILS analytical solution over a cross section $y=30$ m (blue line), temperature correction distributed along $y=30$ m on top of ILS base (red line) and hybrid temperature derived from combination of ILS base and correction (green line).}
    \label{fig:ils_25boreholes_profile}
\end{figure}

\begin{figure}[p]
    \centering
    \begin{subfigure}{\textwidth}
        \centering
        \includegraphics[width=0.95\textwidth]{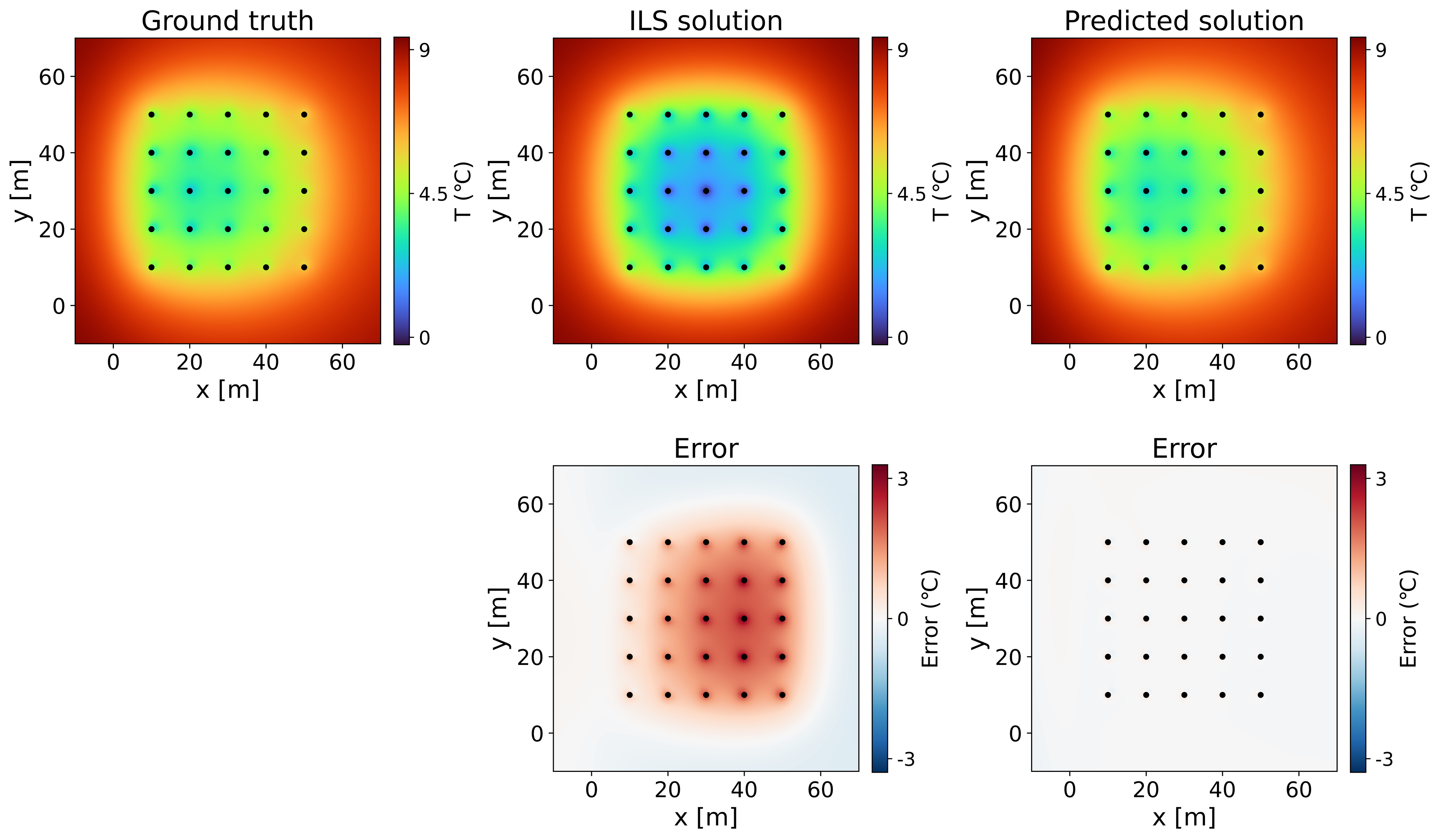}
        \caption{Scenario $1$}
    \end{subfigure}
    \vspace{5mm}  
    \begin{subfigure}{\textwidth}
        \centering
        \includegraphics[width=0.95\textwidth]{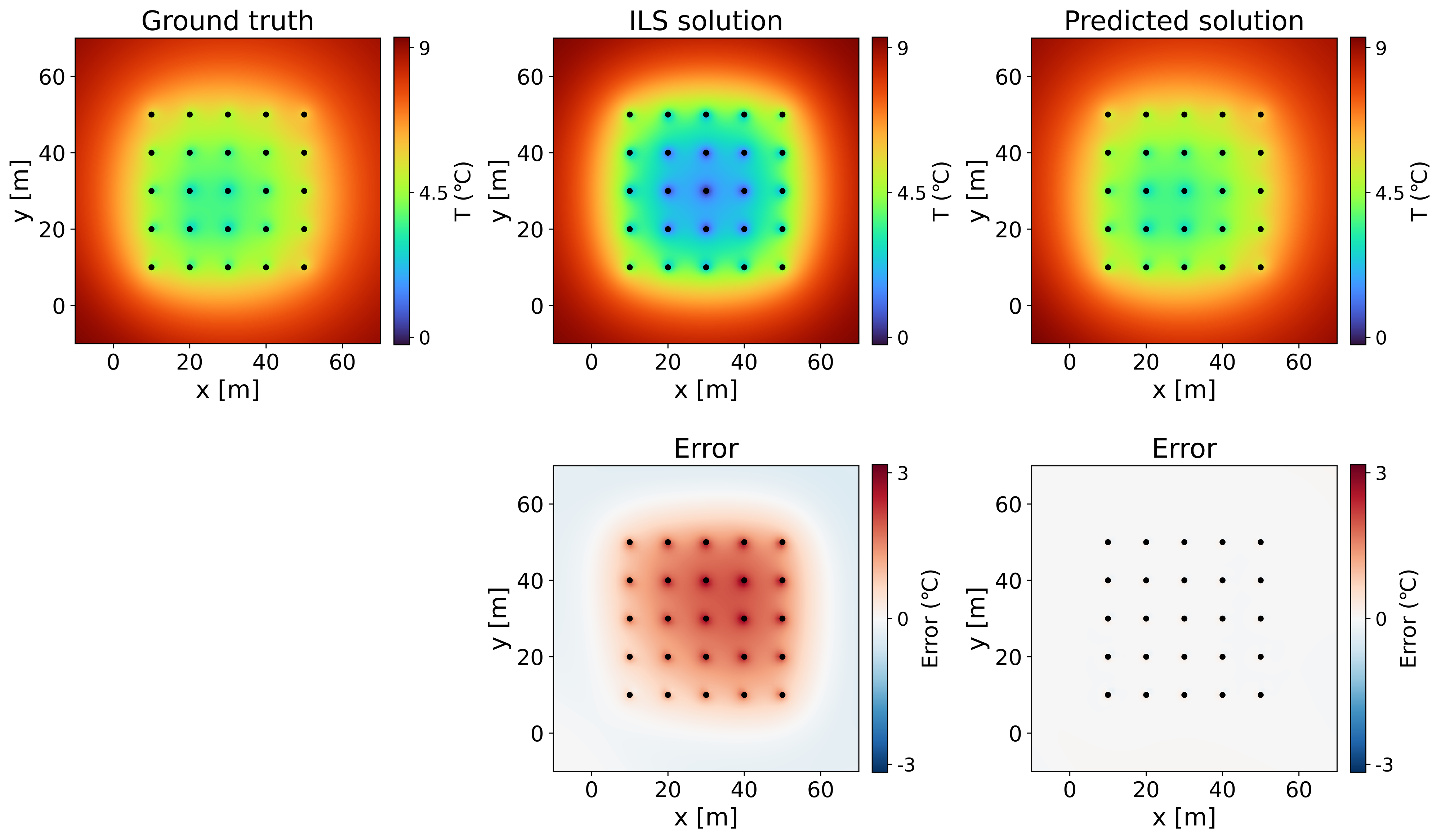}
        \caption{Scenario $2$}
    \end{subfigure}
    \caption{Results in December of the 15th year for (a) Scenario 1 and (b) Scenario 2. For each panel: the first row, soil temperature distributions derived from the finite difference solver with mesh size $h=1$m (Ground truth), global ILS analytical formula (ILS solution) and hybrid-PINN corrector based on ILS solution (Predicted solution); the second row, the difference distributions of ILS solution and predicted solution with respect to the ground truth.}
    \label{fig:ils_x_15_open_error}
\end{figure}

The compared distributions by numerical methods, analytical models and hybrid correctors are shown in Fig. \ref{fig:ils_x_15_open_error}. For brevity, we only present the results for December of the 15th year in Fig. \ref{fig:ils_x_15_open_error} and provide those for other selected time steps in the Appendix. The ILS errors in Fig. \ref{fig:ils_x_15_open_error} exhibit a structured asymmetry along the conductivity gradient direction, and the corrected solution largely compensates for the discrepancy introduced by the soil heterogeneity in both shape and amplitude. Figure \ref{fig:ils_25boreholes_profile} illustrates the role of the neural correction in heterogeneous soil temperature simulation in Decembers of years 5, 15 and 30. The ILS solution captures the dominant structure of the temperature field, and the learned correction provides localized adjustments around the boreholes and captures the spatially nonuniform error induced by soil heterogeneity. The neural correction also helps stabilize the growing analytical error over time.
\begin{table}[!htbp]
\centering
\caption{$L^2$ and $L^{\infty}$ relative error comparison among the global ILS solutions, local ILS solutions and hybrid-PINN predictions in Decembers of the $5$th, $15$th and $30$th years for Scenarios 1 and 2.}
\label{tab:ils_x_15_open_relative_errors}
\footnotesize
\setlength{\tabcolsep}{2.5pt}

\begin{tabular}{c c cc cc cc}
\toprule
\multirow{2}{*}{Scenario}
& \multirow{2}{*}{Month}
& \multicolumn{2}{c}{Global ILS}
& \multicolumn{2}{c}{Local ILS}
& \multicolumn{2}{c}{Predicted solution} \\
\cmidrule(lr){3-4}
\cmidrule(lr){5-6}
\cmidrule(lr){7-8}
&
& $L^2$ & $L^{\infty}$
& $L^2$ & $L^{\infty}$
& $L^2$ & $L^{\infty}$ \\
\midrule

\multirow{3}{*}{1}
& 60
& $3.1876\mathrm{e}{-2}$ & $1.8523\mathrm{e}{-1}$
& $9.9915\mathrm{e}{-3}$ & $5.0113\mathrm{e}{-2}$
& $6.0353\mathrm{e}{-3}$ & $3.2807\mathrm{e}{-2}$ \\

& 180
& $1.2711\mathrm{e}{-1}$ & $3.5581\mathrm{e}{-1}$
& $2.5586\mathrm{e}{-2}$ & $5.2715\mathrm{e}{-2}$
& $6.0718\mathrm{e}{-3}$ & $3.3899\mathrm{e}{-2}$ \\

& 360
& $3.2482\mathrm{e}{-1}$ & $6.0071\mathrm{e}{-1}$
& $3.9225\mathrm{e}{-2}$ & $6.6342\mathrm{e}{-2}$
& $1.6855\mathrm{e}{-2}$ & $3.3604\mathrm{e}{-2}$ \\

\midrule

\multirow{3}{*}{2}
& 60
& $3.1300\mathrm{e}{-2}$ & $1.7519\mathrm{e}{-1}$
& $8.1848\mathrm{e}{-3}$ & $5.0269\mathrm{e}{-2}$
& $4.0250\mathrm{e}{-3}$ & $2.8349\mathrm{e}{-2}$ \\

& 180
& $1.2855\mathrm{e}{-1}$ & $3.4183\mathrm{e}{-1}$
& $1.6081\mathrm{e}{-2}$ & $5.1533\mathrm{e}{-2}$
& $4.2203\mathrm{e}{-3}$ & $3.0135\mathrm{e}{-2}$ \\

& 360
& $3.3213\mathrm{e}{-1}$ & $5.7402\mathrm{e}{-1}$
& $2.1162\mathrm{e}{-2}$ & $5.6521\mathrm{e}{-2}$
& $7.6073\mathrm{e}{-3}$ & $3.0417\mathrm{e}{-2}$ \\

\bottomrule
\end{tabular}
\end{table}

\begin{figure}[!htbp]
    \centering
    \includegraphics[width=0.95\textwidth]{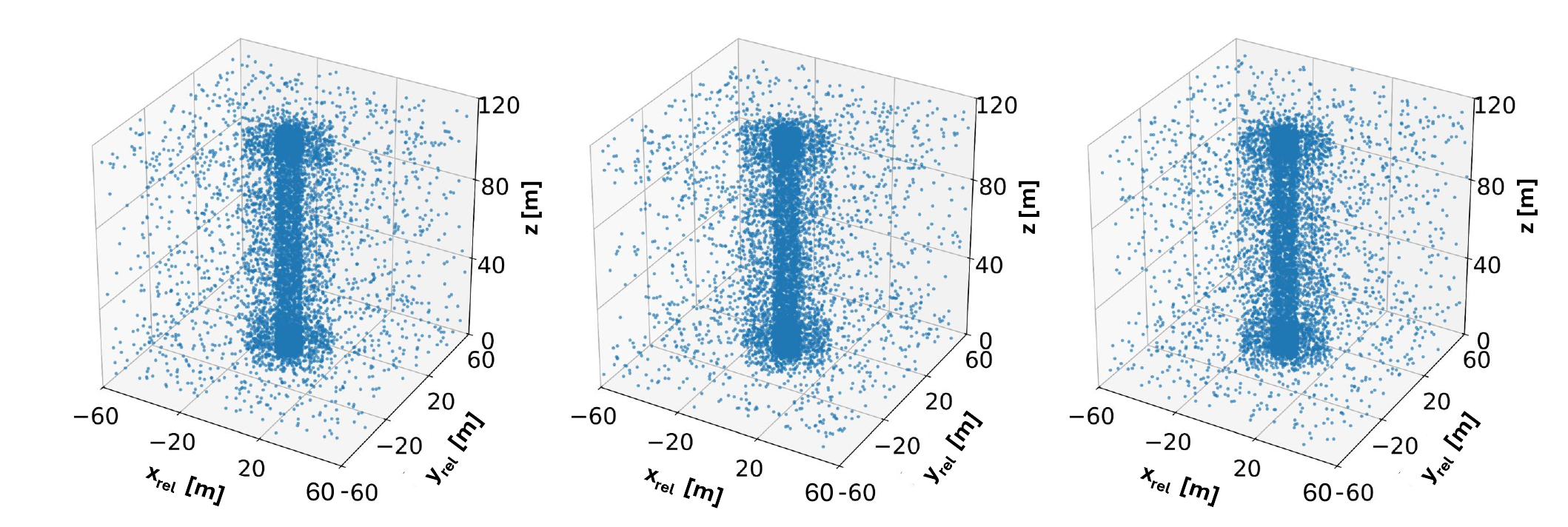}
    \caption{Visualization of PDE collocation points sampled for Examples 2 and 3 in month 1, 60 and 360 (from left to right). }
    \label{fig:fls_sampling}
\end{figure}

\paragraph{Example 2.(FLS-based Model)}
In this example, the same simulation period, BHE arrangement and input sampling are applied as those in Example 1. Here, we consider the computational domain $\left[-10,70\right]\mathrm{m}\times\left[-10,70\right]\mathrm{m}\times\left[0,120\right]\mathrm{m}$ and model each BHE as a vertical line source with zero radius starting at depth $D=10$m and extending to $D+L=110$m. The vertical coordinates of training points are sampled from the source endpoint regions (\(z\in\left[4,16\right]\bigcup \left[104,116\right]\mathrm{m})\), the active source segment (\(z\in\left[10,110\right]\mathrm{m}\)) and the complete vertical plane with probabilities $\beta_0=0.45$, $\beta_1=0.35$ and $\beta_2=0.2$, respectively. We present the PDE collocation point sampling in the three-dimensional space with line source at the horizontal origin in Figure \ref{fig:fls_sampling}. 

\begin{figure}[p]
    \centering
    \begin{subfigure}{\textwidth}
        \centering
        \includegraphics[width=0.95\textwidth]{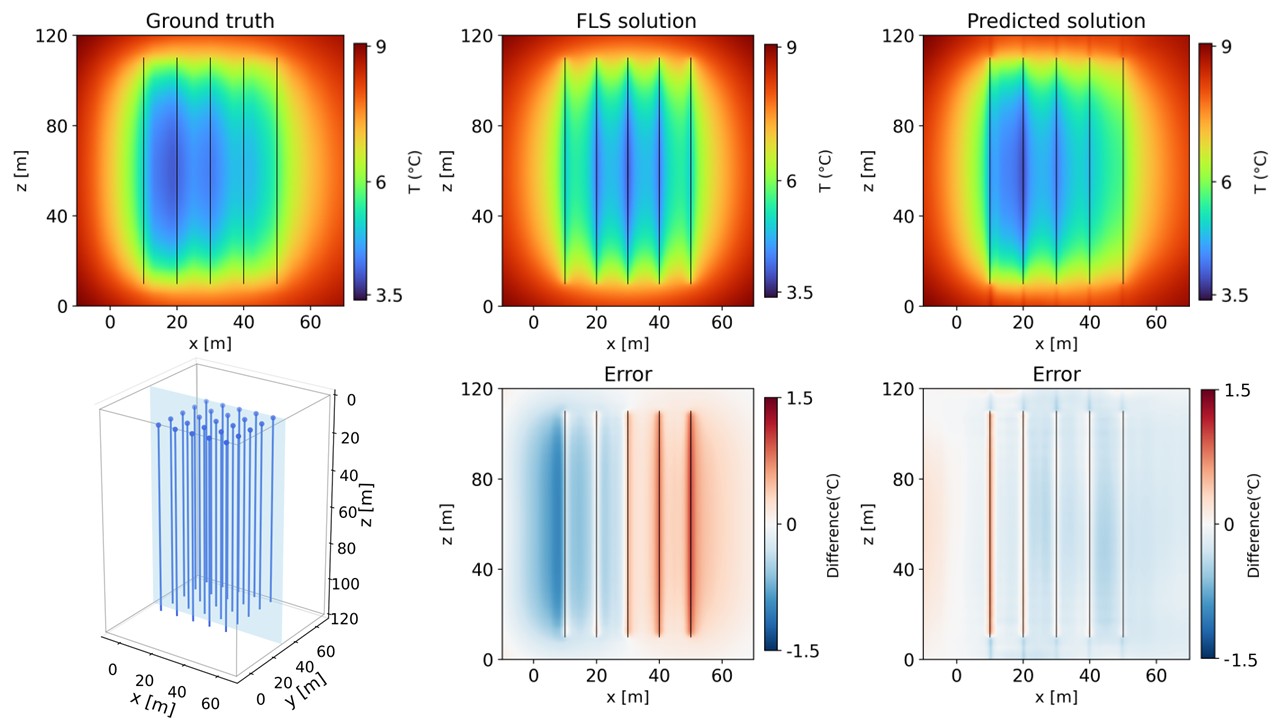}
        \caption{Scenario $1$, evaluated on the vertical cross section at $y=30$ m.}
    \end{subfigure}
    \vspace{5mm}
    \begin{subfigure}{\textwidth}
        \centering
        \includegraphics[width=0.95\textwidth]{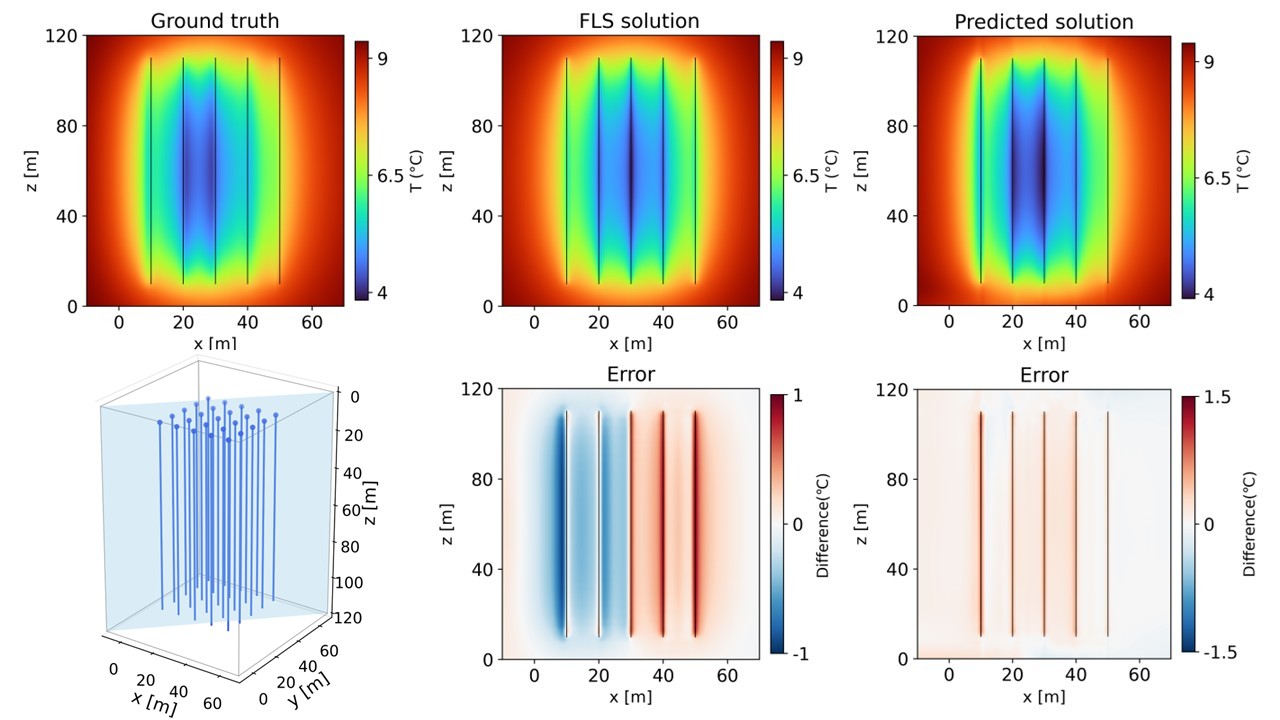}
        \caption{Scenario $2$, evaluated on the vertical cross section at $y=x$.}
    \end{subfigure}
    \caption{Results in December of the 15th year for (a) Scenario 1 and (b) Scenario 2. For each panel: the first row, soil temperature distributions derived from the finite difference solver with mesh size $h=1$ m (Ground truth), global FLS analytical formula (FLS solution) and hybrid-PINN corrector based on the FLS solution (Predicted solution); the second row, the difference distributions of the FLS solution and predicted solution with respect to the ground truth. For Scenario 2, the horizontal coordinate is labeled by the projection onto the $x$ axis.}
    \label{fig:fls_x_15_open_error}
\end{figure}

Figure \ref{fig:fls_x_15_open_error} provides the representative results tested in December of the 15th year for Scenarios 1 and 2, where the temperature fields are visualized on the \(y=30\) m and $y=x$ cross sections, respectively. Similar to the results of Example 1, the use of a single bulk conductivity leads to spatially paired overestimation and underestimation on the selected cross sections as illustrated in FLS solution. As the mismatch between the local and bulk conductivities increases, larger deviations appear around boreholes further from the reference conductivity location. The hybrid correction largely removes these biases and balances the error magnitude across the three-dimensional computation domain.

The relative errors for all three tested time steps are reported in Table \ref{tab:fls_x_open_relative_errors}. The hybrid prediction consistently reduces the \(L^2\) error relative to both global and local FLS approximations and improves the \(L^\infty\) error at moderate and long simulation times. These outputs demonstrate that the mismatch in the overall behavior and magnitude between three-dimensional FLS model and true solution is mitigated by the neural correction $\widetilde{T}^{\mathrm{FLS}}_{\theta^*,c}$ and its role to correct the heterogeneity-induced bias is fulfilled.

\begin{table}[!htbp]
\centering
\caption{$L^2$ and $L^{\infty}$ relative error comparison among the global FLS solutions, local FLS solutions and hybrid-PINN predictions in Decembers of the $5$th, $15$th and $30$th years for Scenarios 1 and 2, computed over the vertical cross sections at $y=30$ m and $y=x$, respectively.}
\label{tab:fls_x_open_relative_errors}
\footnotesize
\setlength{\tabcolsep}{3.0pt}

\begin{tabular}{c c cc cc cc}
\toprule
\multirow{2}{*}{Scenario}
& \multirow{2}{*}{Month}
& \multicolumn{2}{c}{Global FLS}
& \multicolumn{2}{c}{Local FLS}
& \multicolumn{2}{c}{Predicted solution} \\
\cmidrule(lr){3-4}
\cmidrule(lr){5-6}
\cmidrule(lr){7-8}
&
& $L^2$ & $L^{\infty}$
& $L^2$ & $L^{\infty}$
& $L^2$ & $L^{\infty}$ \\
\midrule

\multirow{3}{*}{1}
& 60
& $2.6404\mathrm{e}{-2}$ & $8.5160\mathrm{e}{-2}$
& $2.3119\mathrm{e}{-2}$ & $9.6849\mathrm{e}{-2}$
& $1.9675\mathrm{e}{-2}$ & $8.7483\mathrm{e}{-2}$ \\

& 180
& $5.9953\mathrm{e}{-2}$ & $1.1768\mathrm{e}{-1}$
& $2.9276\mathrm{e}{-2}$ & $8.8571\mathrm{e}{-2}$
& $2.4528\mathrm{e}{-2}$ & $8.2761\mathrm{e}{-2}$ \\

& 360
& $1.0527\mathrm{e}{-1}$ & $1.7150\mathrm{e}{-1}$
& $4.7142\mathrm{e}{-2}$ & $8.0883\mathrm{e}{-2}$
& $3.5287\mathrm{e}{-2}$ & $8.0621\mathrm{e}{-2}$ \\

\midrule

\multirow{3}{*}{2}
& 60
& $2.3025\mathrm{e}{-2}$ & $8.0344\mathrm{e}{-2}$
& $2.2462\mathrm{e}{-2}$ & $9.2677\mathrm{e}{-2}$
& $1.8683\mathrm{e}{-2}$ & $9.0873\mathrm{e}{-2}$ \\

& 180
& $4.1667\mathrm{e}{-2}$ & $1.0433\mathrm{e}{-1}$
& $2.5269\mathrm{e}{-2}$ & $8.5806\mathrm{e}{-2}$
& $2.1829\mathrm{e}{-2}$ & $8.0854\mathrm{e}{-2}$ \\

& 360
& $6.8374\mathrm{e}{-2}$ & $1.4740\mathrm{e}{-1}$
& $3.4778\mathrm{e}{-2}$ & $8.2350\mathrm{e}{-2}$
& $2.4409\mathrm{e}{-2}$ & $8.0451\mathrm{e}{-2}$ \\

\bottomrule
\end{tabular}
\end{table}

\begin{figure}[t]
    \centering
    \begin{subfigure}{\textwidth}
        \centering
        \includegraphics[width=0.95\textwidth]{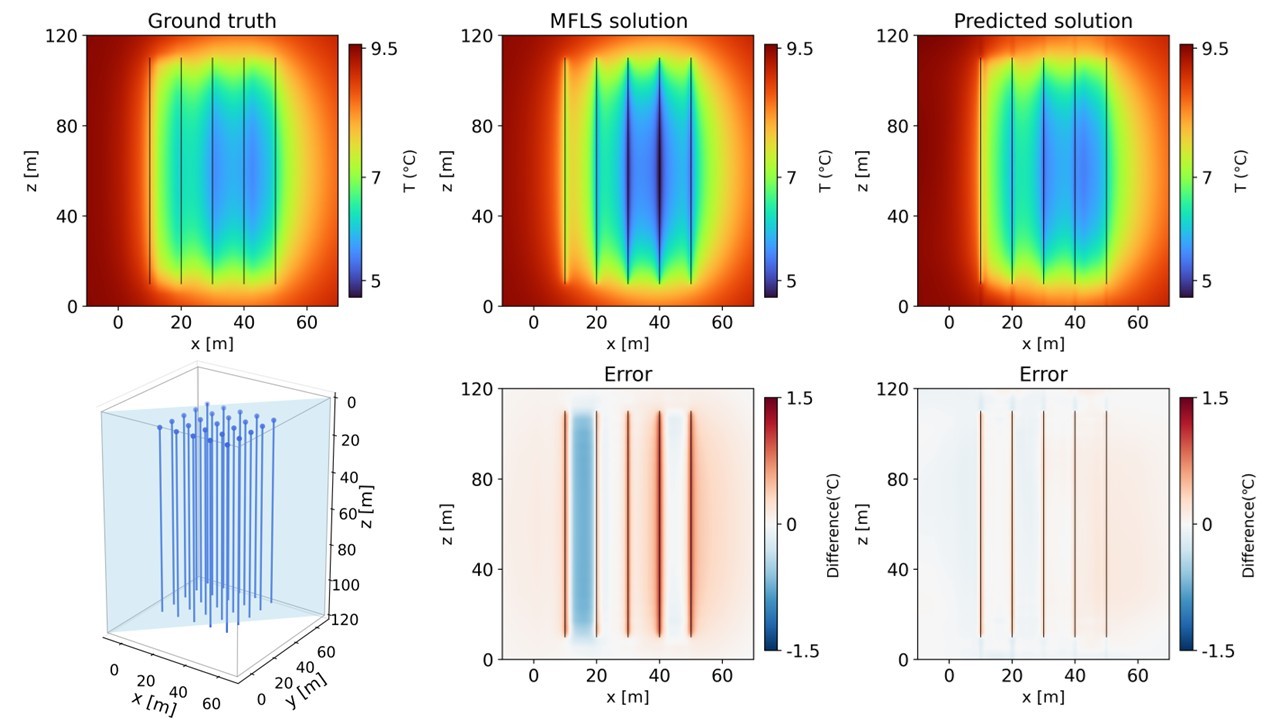}
        \caption{Vertical cross section at $y=x$.}
    \end{subfigure}
    \vspace{5mm}
    \begin{subfigure}{\textwidth}
        \centering
        \includegraphics[width=0.95\textwidth]{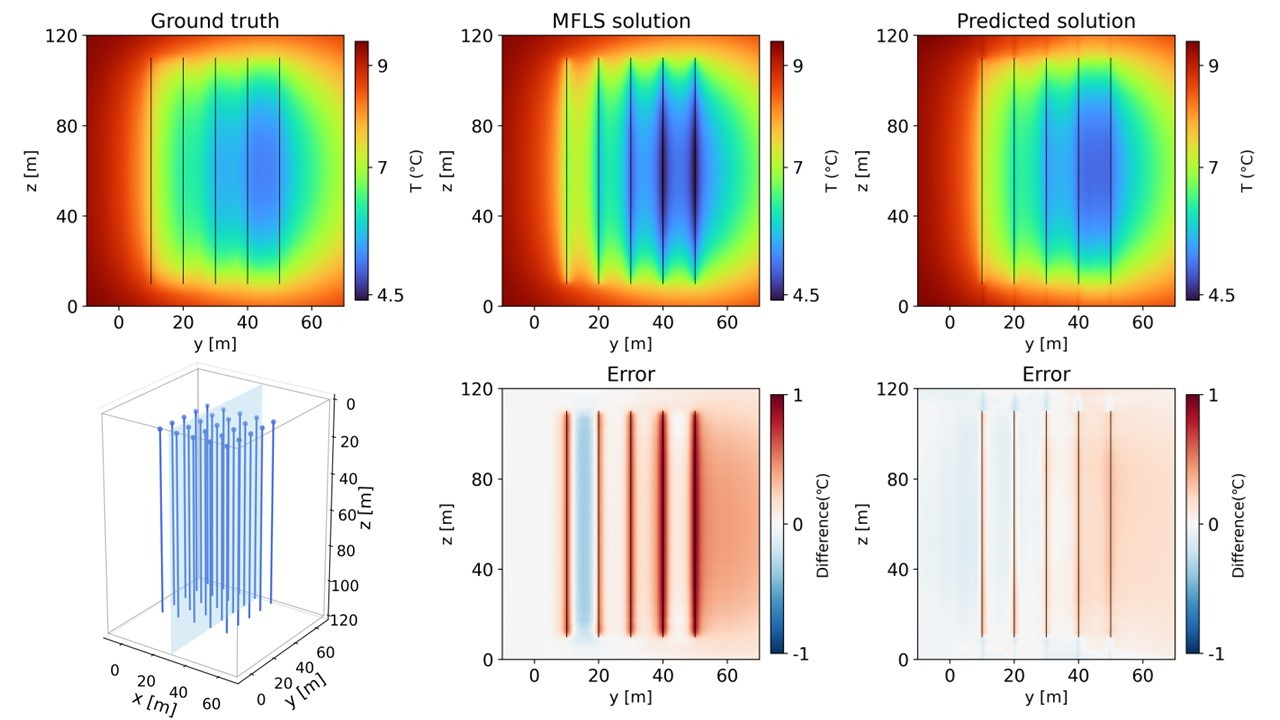}
        \caption{Vertical cross section at $x=30$ m.}
    \end{subfigure}
    \caption{Results in December of the $15$th year for Scenario 2 on (a) the vertical cross section at $y=x$ and (b) the vertical cross section at $x=30$ m. For each panel: the first row, soil temperature distributions derived from the finite difference solver with mesh size $h=1$ m (Ground truth), global MFLS analytical formula (MFLS solution) and hybrid-PINN corrector based on the MFLS solution (Predicted solution); the second row, the difference distributions of the MFLS solution and predicted solution with respect to the ground truth. For panel (a), the horizontal coordinate is labeled by the projection onto the $x$ axis.}
    \label{fig:mfls_x_15_ymid_error}
\end{figure}

\paragraph{Example 3.(MFLS-based Model)}
We employ the same computational setup and training configuration as Example 2 to this case. We choose $\mathrm{\mathbf{v}}=\left(0,2,0\right)$ m/year for training and inference in this test. To examine the effects of conductivity and groundwater advection, we evaluate the predicted temperature fields on two representative vertical sections which are aligned with the conductivity gradient and groundwater flow direction respectively for both scenarios. The simulation results for Scenario 2 are presented in Fig. \ref{fig:mfls_x_15_ymid_error}. The corresponding results for Scenario 1 and the horizontal slices at \(z=60\) m for Scenario 2 are provided in the Appendix. For both cross sections selected along the conductivity gradient and groundwater flow direction, respectively, the learned correction adjusts the magnitude of the temperature response based on the MFLS solution which captures the advection-induced displacement of the temperature field, leading to substantially improved agreement with the numerical reference.  

Table \ref{tab:mfls_x_relative_errors} shows the comparison between the accuracy performance of global MFLS solution, local MFLS solution and predicted approximation on the cross sections $y=30$ m, $y=x$ for Scenarios 1 and 2. Across three selected time steps, the hybrid prediction achieves the lowest overall errors among the three approaches, with its relative $L^2$ error following the analytical temporal trend at a consistently lower level. Notably, the relative \(L^2\) and $L^{\infty}$ errors remain below \(2\%\) and \(6\%\), respectively, for all three evaluated time steps. 

\begin{table}[bth]
\centering
\caption{$L^2$ and $L^{\infty}$ relative error comparison among the global MFLS solutions, local MFLS solutions and hybrid-PINN predictions in Decembers of the $5$th, $15$th, and $30$th years for Scenarios 1 and 2, computed over the vertical cross sections at $y=30$ m and $y=x$, respectively.}
\label{tab:mfls_x_relative_errors}
\footnotesize
\setlength{\tabcolsep}{3.0pt}

\vspace{-0.8em}
\begin{tabular}{c c cc cc cc}
\toprule
\multirow{2}{*}{Scenario}
& \multirow{2}{*}{Month}
& \multicolumn{2}{c}{Global MFLS}
& \multicolumn{2}{c}{Local MFLS}
& \multicolumn{2}{c}{Predicted solution} \\
\cmidrule(lr){3-4}
\cmidrule(lr){5-6}
\cmidrule(lr){7-8}
&
& $L^2$ & $L^{\infty}$
& $L^2$ & $L^{\infty}$
& $L^2$ & $L^{\infty}$ \\
\midrule

\multirow{3}{*}{1}
& 60
& $3.1835\mathrm{e}{-2}$ & $9.5800\mathrm{e}{-2}$
& $3.3631\mathrm{e}{-2}$ & $1.3052\mathrm{e}{-1}$
& $1.2883\mathrm{e}{-2}$ & $5.4947\mathrm{e}{-2}$ \\

& 180
& $3.8989\mathrm{e}{-2}$ & $1.0668\mathrm{e}{-1}$
& $3.8217\mathrm{e}{-2}$ & $1.3185\mathrm{e}{-1}$
& $1.4186\mathrm{e}{-2}$ & $5.8802\mathrm{e}{-2}$ \\

& 360
& $3.9346\mathrm{e}{-2}$ & $1.0401\mathrm{e}{-1}$
& $3.7546\mathrm{e}{-2}$ & $1.2719\mathrm{e}{-1}$
& $1.9900\mathrm{e}{-2}$ & $5.1020\mathrm{e}{-2}$ \\

\midrule

\multirow{3}{*}{2}
& 60
& $3.0866\mathrm{e}{-2}$ & $9.4315\mathrm{e}{-2}$
& $3.2402\mathrm{e}{-2}$ & $1.3027\mathrm{e}{-1}$
& $1.0262\mathrm{e}{-2}$ & $5.5186\mathrm{e}{-2}$ \\

& 180
& $3.9475\mathrm{e}{-2}$ & $1.0962\mathrm{e}{-1}$
& $3.8389\mathrm{e}{-2}$ & $1.2948\mathrm{e}{-1}$
& $1.3497\mathrm{e}{-2}$ & $5.3051\mathrm{e}{-2}$ \\

& 360
& $4.2862\mathrm{e}{-2}$ & $1.0943\mathrm{e}{-1}$
& $4.0260\mathrm{e}{-2}$ & $1.2774\mathrm{e}{-1}$
& $1.5536\mathrm{e}{-2}$ & $5.2131\mathrm{e}{-2}$ \\

\bottomrule
\end{tabular}
\end{table}

\vspace{-2.2em}
\paragraph{Runtime Comparison}
To evaluate the computational benefits of the hybrid PINN model, we compare its inference time against the traditional FDM solver across three-dimensional analytical solution based models at three time instances. The grid spacing for the implicit finite difference solver is set to $4$ m in all three directions and the transient simulation is advanced with a monthly time step. For a fair timing comparison, the hybrid PINN prediction is evaluated on the same three-dimensional grid with the same spatial spacing as the FDM reference. The reported full computation time includes the analytical baseline evaluation and the neural correction superposition over all boreholes and all monthly load increments up to the target time. The comparison for FLS-based model is shown in Table \ref{tab:fls_timing} and for MFLS-based model in Table \ref{tab:mfls_timing}. The hybrid model achieves a substantial computation speed-up compared with traditional numerical solver, which can be particularly beneficial for repeated evaluations where the numerical solver cost becomes prohibitive. Further mesh refinement as used in the accuracy tests would increase the computational cost and amplify the efficiency advantage of the hybrid model. However, the analytical base calculation currently accounts for the majority of the computation cost of the hybrid model, especially for long-term simulations. This component could be further accelerated using efficient computational techniques such as FFT-based convolution and temporal aggregation \cite{heinzel2026optimization}.
\begin{table*}[!htbp]
\centering
\caption{Computation times (in seconds) of the FDM solver and FLS-based hybrid model for tested Scenarios.}
\label{tab:fls_timing}
\footnotesize
\renewcommand{\arraystretch}{1.10}
\setlength{\tabcolsep}{3pt}

\begin{tabular}{
    c c c c c c c
}
\toprule
Scenario
& Month
& {\makecell{FDM}}
& {\makecell{Single NN\\inference}}
& {\makecell{Correction\\superposition}}
& {\makecell{Analytical base\\calculation}}
& {\makecell{Hybrid\\inference}} \\
\midrule

\multirow{3}{*}{1}
& 60
& 3.2563e+3
& 4.1455e-4
& 2.7130e+0
& 1.6553e+1
& 1.9266e+1 \\

& 180
& 3.4348e+3
& 4.8399e-4
& 7.8289e+0
& 4.7147e+1
& 5.4976e+1 \\

& 360
& 3.6993e+3
& 4.9350e-4
& 1.5109e+1
& 8.7967e+1
& 1.0308e+2 \\

\midrule

\multirow{3}{*}{2}
& 60
& 2.6973e+3
& 4.9699e-4
& 2.9729e+0
& 1.7032e+1
& 2.0005e+1 \\

& 180
& 2.8629e+3
& 5.0324e-4
& 8.1385e+0
& 4.8447e+1
& 5.6585e+1 \\

& 360
& 3.1115e+3
& 4.9975e-4
& 1.4934e+1
& 9.0041e+1
& 1.0498e+2 \\

\bottomrule
\end{tabular}
\end{table*}

\vspace{-1em}

\begin{table*}[!htpb]
\centering
\caption{Computation times (in seconds) of the FDM solver and MFLS-based hybrid model for tested Scenarios.}
\label{tab:mfls_timing}
\footnotesize
\renewcommand{\arraystretch}{1.10}
\setlength{\tabcolsep}{3pt}

\begin{tabular}{
    c c c c c c c
}
\toprule
Scenario
& Month
& {\makecell{FDM}}
& {\makecell{Single NN\\inference}}
& {\makecell{Correction\\superposition}}
& {\makecell{Analytical base\\calculation}}
& {\makecell{Hybrid\\inference}} \\
\midrule

\multirow{3}{*}{1}
& 60
& 3.5769e+3
& 5.2584e-4
& 2.7030e+0
& 3.3293e+1
& 3.5996e+1 \\

& 180
& 3.7872e+3
& 3.8970e-4
& 7.5361e+0
& 9.8276e+1
& 1.0581e+2 \\

& 360
& 4.1029e+3
& 5.4626e-4
& 1.4493e+1
& 1.9920e+2
& 2.1370e+2 \\

\midrule

\multirow{3}{*}{2}
& 60
& 4.7611e+3
& 5.6792e-4
& 2.6806e+0
& 3.5057e+1
& 3.7738e+1 \\

& 180
& 4.9882e+3
& 4.9115e-4
& 7.5115e+0
& 1.0238e+2
& 1.0990e+2 \\

& 360
& 5.3364e+3
& 5.2245e-4
& 1.4764e+1
& 1.9290e+2
& 2.0766e+2 \\

\bottomrule
\end{tabular}
\end{table*}

\FloatBarrier
\section{Conclusion}\label{conclusion}
In this paper, we develop a parametric hybrid-PINN method, based on ILS, FLS and MFLS analytical models, for thermal simulation of
borehole heat exchangers, i.e. singular line sources, in heterogeneous subsurface. By decomposing the global soil temperature change into an analytical baseline and a learned correction, we remove the challenge for the network to accurately capture the singularity and avoid the need to approach the delta function using traditional regularization tools. The goal of the correction component is to characterize mismatch induced by soil properties on account of the idealized assumptions of the analytical models on the soil homogeneity. 

A parametric PINN with source-aware features and adaptive sampling is employed to learn a reusable correction over a range of heterogeneous conductivity conditions. The inference efficiency is attained through applying the spatial and temporal superposition principles to the trained universal corrector. Numerical experiments based on the ILS, FLS and MFLS models demonstrate that the proposed hybrid framework consistently improves the accuracy of analytical approximations by recovering the heterogeneity induced discrepancy in the temperature field. The learned correction suppresses the growth of analytical errors over long simulation periods and maintains stable prediction accuracy under both conductive and advective heat transfer regimes. Meanwhile, the hybrid model substantially reduces the computational cost compared with the FDM solver in the tested cases. These results demonstrate that the proposed framework bridges the gap between the efficiency of analytical models and the ability of numerical methods to resolve heterogeneous subsurface physics, providing improved accuracy and faster simulation for long-term multi-BHE problems.

Future research will extend to providing more comprehensive models for BHE simulation by coupling the heat transport inside the borehole to soil temperature response. The fast and differentiable nature of the learned corrector also makes it promising for optimization tasks, including borehole layout design, installation length, load profile planning and long-term operational scheduling. Another future development is to explore frameworks which can solve more complex soil property fields such as layered subsurface and include a broader set of design and operation parameters. As the parametric space becomes higher-dimensional, more expressive learning architectures including neural operators, encoder-decoder structures and generative deep learning models, may be required to construct scalable surrogate models. Incorporating these directions will further improve the proposed methodology, enabling it to serve as a more extensive tool for the efficient simulation and optimization of geothermal energy systems.

\section*{Conflict of Interest}
The authors declare no conflicts of interest.

\section*{Acknowledgement}
\label{sec:Acknowledgement}
The work presented in this paper was supported by the German Federal Ministry of Economics and Climate Protection (BMWK) within the scope of the research project OptiGeoS (01256700/1).

\bibliographystyle{unsrtnat}
\bibliography{references}

\section*{Appendix}
\paragraph{Problems with Dirichlet boundary conditions.}
To explore the feasibility of the proposed physics-informed corrector under the condition that information of boundary temperature is accessible, we apply the model to predict the solution to (\ref{eq:conduction_soil_model}) subject to an initial condition (\ref{total_temperature_initial_condition}) and a constant Dirichlet boundary condition
\begin{align}
    T(\mathbf{x},t)=T_u,\quad \mathbf{x}\in\partial\Omega_s,\ t\in I.
\notag
\end{align}
For the equation (\ref{eq:conduction_soil_change_model}) governing the temperature change $\widetilde{T}$, the equivalent conditions are imposed by (\ref{eq:initial_Ttilde}) and 
\begin{equation}
\widetilde{T}(\mathbf{x},t)=0,\quad \mathbf{x}\in\partial\Omega_s,\ t\in I.
\label{eq:boundary_Ttilde}
\end{equation}
Since the ILS and FLS based corrections mainly consist of the calibration of two physical mismatches: one induced by the imbalance between averaged and variable conductivities and one caused by difference between boundary conditions for infinite and finite domains. Thus, correction equations (\ref{correction_ils}) and (\ref{eq:correction_fls}) fulfill
\begin{align}
    \widetilde{T}_{c}^{m}(\mathbf{x},t)=-
\widetilde{T}^{m}(\mathbf{x},t),
\quad\mathbf{x}\in\partial\Omega_s,\quad t\in I, \quad \mathrm{m}\in\{\mathrm{ILS}, \mathrm{FLS}\}.\notag
\end{align}
The correction for formation property mismatch appears as an interior forcing term in the correction equations, while the correction for boundary temperature is inherited from the difference between the idealized boundary assumptions of ILS/FLS models and the prescribed boundary conditions of the original problem. For clarity, we consider a homogeneous finite-domain solution $T_{h}$ with constant conductivity $\lambda_0$. Then the boundary correction can be defined for $\mathrm{m}\in\{\mathrm{ILS}, \mathrm{FLS}\}$ as
\begin{equation}
B(\mathbf{x},t)
=T_h(\mathbf{x},t)-\left(\widetilde{T}^{m}(\mathbf{x},t)+T_u\right)
=\widetilde{T}_h(\mathbf{x},t)-\widetilde{T}^{m}(\mathbf{x},t)\notag
\end{equation}
with $\widetilde{T}_h$ denoting the change of the reference temperature solution, which satisfies a homogeneous heat equation,
\begin{equation}
\rho_s C_s \frac{\partial B}{\partial t}
-
\lambda_0 \Delta B
=
0,
\qquad
\mathbf{x}\in \Omega_s,\quad t\in I,
\label{eq:boundary_correction_pde}
\end{equation}
with the inherited boundary condition
\begin{equation}
B(\mathbf{x},t)=-\widetilde{T}^m(\mathbf{x},t),
\qquad
\mathbf{x}\in\partial\Omega_s,\quad t\in I.
\label{eq:boundary_correction_bc}
\end{equation}
Thus, the boundary correction is a well-defined component governed by the inconsistency between the boundary conditions. 

\paragraph{Boundary condition enforcement.}
In a vanilla PINN formulation, the Dirichlet boundary conditions are implemented using additional penalization terms as a contribution to the loss function. However, this treatment is particularly inefficient for the present boundary correction problem (\ref{eq:boundary_correction_pde}) with (\ref{eq:boundary_correction_bc}). This difficulty originates from the fact that the line source response is strongly localized near the borehole at early times, while the thermal perturbation gradually propagates across the domain and interacts with the external boundary at later times. This produces a pronounced imbalance in both spatial locations and temporal scales, making the approximation sensitive to the sampling of residual and boundary collocation points. Moreover, statically assigned weights to different loss terms is generally inadequate to accommodate the continuously evolving distribution of the correction field, making the optimization sensitive to hyperparameter tuning and resulting in substantial computational cost. Therefore, for the FLS-based correction equation, we adopt a hard-constrained strategy based on a distance function to match the specified correction boundary values.

Inspired by \cite{Berrone2023DirichletPINN}, we add a non-trainable layer $N$ at the end of the neural network to modify the previous output $\widetilde{T}_{\theta,c}^{\mathrm{FLS}}$ by
\begin{align}
    N\widetilde{T}_{\theta,c}^{\mathrm{FLS}}=-\bar{\widetilde{T}}^{\mathrm{FLS}}+\phi\widetilde{T}_{\theta,c}^{\mathrm{FLS}}\notag
\end{align}
to exactly satisfy $N\widetilde{T}_{\theta,c}^{\mathrm{FLS}}\left(\mathbf{x},t\right)=-\widetilde{T}^{\mathrm{FLS}}\left(\mathbf{x},t\right)$ on $\mathbf{x}\in\partial \Omega_s$, where $\bar{\widetilde{T}}^{\mathrm{FLS}}$ is a continuous extension of $\widetilde{T}^{\mathrm{FLS}}$ into $\Omega_s$, i.e., $\bar{\widetilde{T}}^{\mathrm{FLS}}|_{\partial\Omega_s}=\widetilde{T}^{\mathrm{FLS}}$ and $\phi$ is an approximate distance function to $\partial \Omega_s$ which satisfies $\phi\left(\mathbf{x}\right)=0$ only for $\mathbf{x}\in\partial\Omega_s$ and $\phi\left(\mathbf{x}\right)>0$ for $\mathbf{x}\in\Omega_s$. The training of the FLS-based model is to minimize the loss corresponding to $N\widetilde{T}_{\theta,c}^{\mathrm{FLS}}$. For the sake of simplicity, we consider $\Omega_s=\left(x_{\min},x_{\max}\right)\times\left(y_{\min},y_{\max}\right)\times\left(z_{\min},z_{\max}\right)$ here, although approximate distance function can be generalized for more complex geometries in three-dimensional space. For the cuboidal domain, \(\phi\) is assembled from the normalized distances to the six boundary faces
\[
\hat{d}_1=\frac{x-x_{\min}}{L_x},\quad \hat{d}_2=\frac{x_{\max}-x}{L_x},\quad
\hat{d}_3=\frac{y-y_{\min}}{L_y},\quad \hat{d}_4=\frac{y_{\max}-y}{L_y},\quad
\hat{d}_5=\frac{z-z_{\min}}{L_z},\quad \hat{d}_6=\frac{z_{\max}-z}{L_z},
\]
where $L_x=x_{\max}-x_{\min}$, $L_y=y_{\max}-y_{\min}$ and $L_z=z_{\max}-z_{\min}$. The approximation distance function is then defined using the normalized form of order $p$
\[
\phi(\mathbf{x})
=
\left(
\sum_{i=1}^{6} \hat{d}_i(\mathbf{x})^{-p}
\right)^{-1/p},
\qquad p\ge 1 .
\]
Since $\widetilde{T}^{\mathrm{FLS}}$ is well-defined and continuous in $\Omega_s\setminus\{\left(0,0,z\right)\mid z\in \left[D,D+L\right]\}$ for $t>0$, we select $\bar{\widetilde{T}}^{\mathrm{FLS}}\left(\mathbf{x},t\right)=\widetilde{T}^{\mathrm{FLS}}\left(\mathbf{x},t\right)$, $\mathbf{x}\in\bar{\Omega}_s$. The transfinite interpolation can also be employed to construct the extended boundary function \cite{Berrone2023DirichletPINN}. 

\paragraph{Location indicator augmentation.} For the ILS-based model with no hard-constrained boundary transformation used, we observe that the boundary condition was less accurately fitted in the
direction orthogonal to the conductivity gradient. This suggests that learning the solution under both absolute and relative coordinate systems using a single neural network structure requires additional efforts. We therefore introduce normalized global coordinate indicators
\begin{align}
    \hat{x}=\hat{d}_1,\qquad \hat{y}=\hat{d}_3\notag
\end{align}
to encode the position of each data or collocation point relative to the computational domain boundary. These indicators are also used in the FLS correctors with
\begin{align}
    \hat{z}=\hat{d}_5\notag
\end{align}
added to better represent the correction in the interior induced by the boundary condition and to better approach the prescribed boundary behavior near $\partial\Omega_s$. That is to say the relative positions with regard to the borehole which indicates the local geometry and to the boundary interfaces which reflects the global characteristics are both incorporated as feature inputs to the neural net model.

\paragraph{Selection of training points.} We add collocation points to evaluate the boundary condition residual to the training points set for ILS-based model $\mathcal{S}_{\mathrm{st}}^{\mathrm{ILS}}$ to obtain
\begin{align}
    \mathcal{S}^{'\mathrm{ILS}}_{\mathrm{st}}=\mathcal{S}_{\mathrm{pde}}^{\mathrm{ILS}}\bigcup\mathcal{S}_{\mathrm{ic}}^{\mathrm{ILS}}\bigcup\mathcal{S}^{\mathrm{ILS}}_{\mathrm{bc}}\bigcup\mathcal{S}^{\mathrm{ILS}}_{\mathrm{data}},\notag
\end{align}
where
\begin{align}
\mathcal{S}^{\mathrm{ILS}}_{\mathrm{bc}}=\Bigl\{\left(x_i,y_i,t_i\right)|\left(x_i,y_i\right)\sim\mathcal{U}
\!\left(
\partial\Omega_s
\right),t_i\sim\mathcal{U}\left(I\right)\Bigr\}_{i=1}^{N_{\mathrm{bc}}}.\notag
\end{align}
FLS-based model with hard-enforced boundary conditions can use the previously defined training set $\mathcal{S}^{\mathrm{FLS}/\mathrm{MFLS}}_{\mathrm{st}}$. Thus the loss function for ILS-based model is adjusted as
\begin{equation}
\mathcal{L}^{\mathrm{ILS}}
=
w_{\mathrm{data}}^{\mathrm{ILS}}\mathcal{L}_{\mathrm{data}}^{\mathrm{ILS}}
+
w_{\mathrm{pde}}^{\mathrm{ILS}}\mathcal{L}_{\mathrm{pde}}^{\mathrm{ILS}}
+
w_{\mathrm{ic}}^{\mathrm{ILS}}\mathcal{L}_{\mathrm{ic}}^{\mathrm{ILS}}
+
w_{\mathrm{bc}}^{\mathrm{ILS}}\mathcal{L}_{\mathrm{bc}}^{\mathrm{ILS}},\notag
\end{equation}
where
\begin{equation}
\mathcal{L}_{\mathrm{data}}^{\mathrm{ILS}}
=
\frac{1}{N_{p,1}N_{\mathrm{data}}}
\sum_{j=1}^{N_{p,1}}
\sum_{i=1}^{N_{\mathrm{data}}}
\left|
\frac{
\widetilde{T}^{\mathrm{ILS}}_{\theta,c}(x_i,y_i,t_i,r_i,\hat{x}_i,\hat{y}_i;\boldsymbol{\mu}^{\left(j\right)})
-
\widetilde{T}^{\mathrm{ILS}}_{\mathrm{data},c}\left(x_i,y_i,t_i;\boldsymbol{\mu}^{\left(j\right)}\right)}{C_{\mathrm{scale}}}\right|^2,\notag
\end{equation}
\begin{equation}
\mathcal{L}_{\mathrm{pde}}^{\mathrm{ILS}}
=
\frac{1}{N_{p,2}N_{\mathrm{pde}}}
\sum_{j=1}^{N_{p,2}}
\sum_{i=1}^{N_{\mathrm{pde}}}
\left|\frac{\left[\rho_sC_s\frac{\partial \widetilde{T}^{\mathrm{ILS}}_{\theta,c}}{\partial t}-\nabla\cdot\left(\lambda^{\left(j\right)}\nabla\widetilde{T}^{\mathrm{ILS}}_{\theta,c}\right)-\nabla\cdot\left(\left(\lambda^{\left(j\right)}-\lambda_0\right)\nabla\widetilde{T}^{\mathrm{ILS}}\right)\right]\left(x_i,y_i,t_i,r_i,\hat{x}_i,\hat{y}_i;\boldsymbol{\mu}^{\left(j\right)}\right)}{R_{\mathrm{scale}}}\right|^2 ,\notag
\end{equation}
\begin{equation}
\mathcal{L}_{\mathrm{ic}}^{\mathrm{ILS}}
=
\frac{1}{N_{p,3}N_{\mathrm{ic}}}
\sum_{j=1}^{N_{p,3}}
\sum_{i=1}^{N_{\mathrm{ic}}}
\left|
\frac{
\widetilde{T}_{\theta,c}^{\mathrm{ILS}}\left(x_i,y_i,0,r_i,\hat{x}_i,\hat{y}_i;\boldsymbol{\mu}^{\left(j\right)}\right)
}{C_{\mathrm{scale}}}\right|^2 ,\notag
\end{equation}
and
\begin{equation}
\mathcal{L}_{\mathrm{bc}}^{\mathrm{ILS}}
=
\frac{1}{N_{p,4}N_{\mathrm{bc}}}
\sum_{j=1}^{N_{p,4}}
\sum_{i=1}^{N_{\mathrm{bc}}}
\left|
\frac{
\widetilde{T}^{\mathrm{ILS}}_{\theta,c}(x_i,y_i,t_i,r_i,\hat{x}_i,\hat{y}_i;\boldsymbol{\mu}^{\left(j\right)})
+
\widetilde{T}^{\mathrm{ILS}}\left(x_i,y_i,t_i;\boldsymbol{\mu}^{\left(j\right)}\right)
}
{C_{\mathrm{scale}}}
\right|^2\notag
\end{equation}
with predefined normalization scales $C_{\mathrm{scale}}$, $R_{\mathrm{scale}}$, $\lambda^{\left(j\right)}=\lambda_B^{\left(j\right)}+\lambda_x^{\left(j\right)}x+\lambda_y^{\left(j\right)}y$, $ r_i=\sqrt{x_i^2+y_i^2}$, and $\hat{x}_i=\frac{x_i-x_{\min}}{x_{\max}-x_{\min}}$, $\hat{y}_i=\frac{y_i-y_{\min}}{y_{\max}-y_{\min}}$. Similarly, the loss function for FLS-based model is reformulated as
\begin{equation}
\mathcal{L}^{\mathrm{FLS}}
=
w_{\mathrm{data}}^{\mathrm{FLS}}\mathcal{L}_{\mathrm{data}}^{\mathrm{FLS}}
+
w_{\mathrm{pde}}^{\mathrm{FLS}}\mathcal{L}_{\mathrm{pde}}^{\mathrm{FLS}}
+
w_{\mathrm{ic}}^{\mathrm{FLS}}\mathcal{L}_{\mathrm{ic}}^{\mathrm{FLS}},\notag
\end{equation}
where
\begin{equation}
\mathcal{L}_{\mathrm{data}}^{\mathrm{FLS}}
=
\frac{1}{N_{p,1}N_{\mathrm{data}}}
\sum_{j=1}^{N_{p,1}}
\sum_{i=1}^{N_{\mathrm{data}}}
\left|
\frac{
\widetilde{T}^{\mathrm{FLS}}_{\theta,c}(x_i,y_i,z_i,t_i,r_i,\hat{x}_i,\hat{y}_i,\hat{z}_i;\boldsymbol{\mu}^{\left(j\right)})
-
\widetilde{T}^{\mathrm{FLS}}_{\mathrm{data},c}\left(x_i,y_i,z_i,t_i;\boldsymbol{\mu}^{\left(j\right)}\right)}{C_{\mathrm{scale}}}\right|^2,\notag
\end{equation}
\begin{align}
\mathcal{L}_{\mathrm{pde}}^{\mathrm{FLS}}
=&
\frac{1}{N_{p,2}N_{\mathrm{pde}}}
\sum_{j=1}^{N_{p,2}}
\sum_{i=1}^{N_{\mathrm{pde}}}\notag\\
&\left|\frac{\left[\rho_sC_s\frac{\partial N\widetilde{T}^{\mathrm{FLS}}_{\theta,c}}{\partial t}-\nabla\cdot\left(\lambda^{\left(j\right)}\nabla N\widetilde{T}^{\mathrm{FLS}}_{\theta,c}\right)-\nabla\cdot\left(\left(\lambda^{\left(j\right)}-\lambda_0\right)\nabla\widetilde{T}^{\mathrm{FLS}}\right)\right]\left(x_i,y_i,z_i,t_i,r_i,\hat{x}_i,\hat{y}_i,\hat{z}_i;\boldsymbol{\mu}^{\left(j\right)}\right)}{R_{\mathrm{scale}}}\right|^2 ,\notag
\end{align}
and
\begin{equation}
\mathcal{L}_{\mathrm{ic}}^{\mathrm{FLS}}
=
\frac{1}{N_{p,3}N_{\mathrm{ic}}}
\sum_{j=1}^{N_{p,3}}
\sum_{i=1}^{N_{\mathrm{ic}}}
\left|
\frac{
N\widetilde{T}_{\theta,c}^{\mathrm{FLS}}\left(x_i,y_i,z_i,0,r_i,\hat{x}_i,\hat{y}_i,\hat{z}_i;\boldsymbol{\mu}^{\left(j\right)}\right)
}{C_{\mathrm{scale}}}\right|^2 ,\notag
\end{equation}
with $
\hat{z}_i=\frac{z_i-z_{\min}}{z_{\max}-z_{\min}}$.
Applying the spatial and temporal superposition principle to the reusable PINN corrector, the approximated temperature by ILS and FLS based models is formulated as
\begin{align}    &T^m_{\theta^*}\left(\mathbf{x},t;\lambda\left(x,y\right)\right)
    =T_{u}\notag+\sum_{k=1}^K\sum_{l=0}^{N_t-1}\left(\widetilde{T}^m_{k,l}\left(\mathbf{x},t;\lambda_0\right)+ \left(q_{k,l}-q_{k,l-1}\right)\widetilde{T}^m_{\theta^*,c}\left(x-x_{\mathrm{BHE}}^k,y-y_{\mathrm{BHE}}^k,z,t-t_l,r,\hat{x},\hat{y},\hat{z};\boldsymbol{\mu}_k\right)\right).\notag
\end{align}

\paragraph{Case study.} We evaluate the models' performance under the same settings as in Examples 1 and 2 in Section \ref{case_study}. ILS and FLS based models are able to recover the late time temperature field with relatively high accuracy under the Dirichlet boundary condition. As displayed in Tables \ref{tab:ils_dirichlet_relative_errors} and \ref{tab:fls_dirichlet_relative_errors}, the $L^2$ relative errors at selected time steps are on the order of $10^{-3}$ and $10^{-2}$ for the ILS-based and FLS-based models, respectively, and a more significant accuracy advantage over the analytical models at later stages of the simulation is demonstrated.
\begin{table}[!htbp]
\centering
\caption{$L^2$ and $L^{\infty}$ relative errors of the global ILS solutions, local ILS solutions and hybrid-PINN predictions in Decembers of years $5$, $15$ and $30$ for Scenarios 1 and 2 under Dirichlet boundary conditions.}
\label{tab:ils_dirichlet_relative_errors}
\footnotesize
\setlength{\tabcolsep}{2.5pt}

\begin{tabular}{c c cc cc cc}
\toprule
\multirow{2}{*}{Scenario}
& \multirow{2}{*}{Month}
& \multicolumn{2}{c}{Global ILS}
& \multicolumn{2}{c}{Local ILS}
& \multicolumn{2}{c}{Predicted solution} \\
\cmidrule(lr){3-4}
\cmidrule(lr){5-6}
\cmidrule(lr){7-8}
&
& $L^2$ & $L^{\infty}$
& $L^2$ & $L^{\infty}$
& $L^2$ & $L^{\infty}$ \\
\midrule

\multirow{3}{*}{1}
& 60
& $2.2994\mathrm{e}{-2}$ & $1.2238\mathrm{e}{-1}$
& $1.7270\mathrm{e}{-2}$ & $5.1837\mathrm{e}{-2}$
& $5.3270\mathrm{e}{-3}$ & $5.5080\mathrm{e}{-2}$ \\

& 180
& $1.1436\mathrm{e}{-1}$ & $1.5807\mathrm{e}{-1}$
& $1.0582\mathrm{e}{-1}$ & $1.6973\mathrm{e}{-1}$
& $6.3225\mathrm{e}{-3}$ & $5.7884\mathrm{e}{-2}$ \\

& 360
& $2.5428\mathrm{e}{-1}$ & $2.8531\mathrm{e}{-1}$
& $2.5348\mathrm{e}{-1}$ & $3.1442\mathrm{e}{-1}$
& $7.1412\mathrm{e}{-3}$ & $5.3680\mathrm{e}{-2}$ \\

\midrule

\multirow{3}{*}{2}
& 60
& $2.0424\mathrm{e}{-2}$ & $1.1411\mathrm{e}{-1}$
& $1.7429\mathrm{e}{-2}$ & $5.0191\mathrm{e}{-2}$
& $4.9878\mathrm{e}{-3}$ & $5.5785\mathrm{e}{-2}$ \\

& 180
& $1.1324\mathrm{e}{-1}$ & $1.5807\mathrm{e}{-1}$
& $1.0956\mathrm{e}{-1}$ & $1.6373\mathrm{e}{-1}$
& $5.4639\mathrm{e}{-3}$ & $5.9050\mathrm{e}{-2}$ \\

& 360
& $2.5934\mathrm{e}{-1}$ & $2.8531\mathrm{e}{-1}$
& $2.5968\mathrm{e}{-1}$ & $2.9941\mathrm{e}{-1}$
& $6.6568\mathrm{e}{-3}$ & $5.6409\mathrm{e}{-2}$ \\

\bottomrule
\end{tabular}
\end{table}

\begin{figure}[!htbp]
    \centering
    \includegraphics[width=0.95\textwidth]{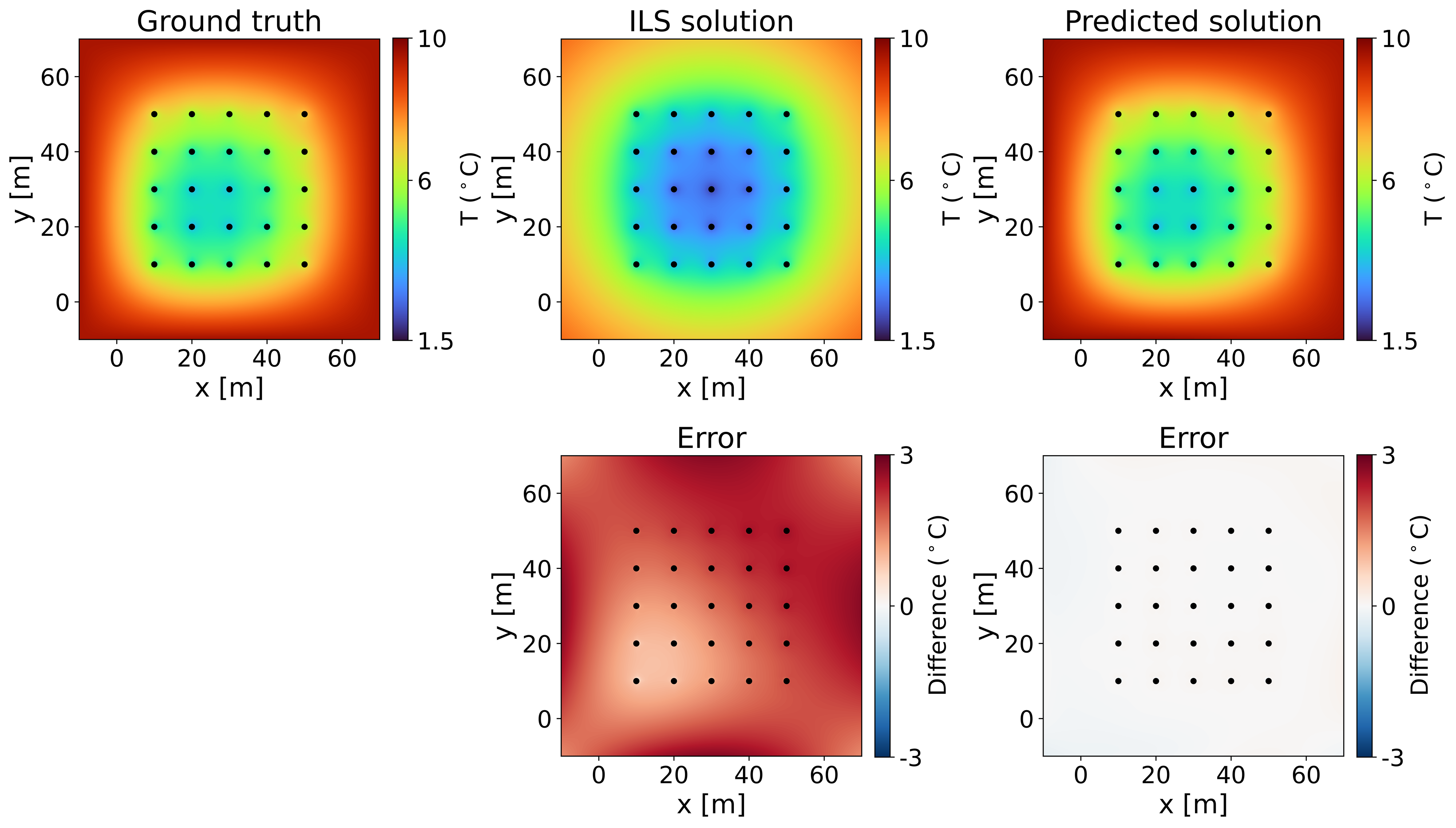}
    \caption{The first row: temperature distributions in December of year 30 for Scenario 2 under Dirichlet boundary condition, derived from the finite difference solver of mesh size $h=1$m (Ground truth), ILS analytical formula (ILS solution) and hybrid-PINN corrector (Predicted solution); the second row: the difference distributions of ILS solution and predicted solution with respect to the ground truth.}
    \label{fig:ils_xy_30_error}
\end{figure}

\begin{figure}[htbp]
    \centering
    \includegraphics[width=0.95\textwidth]{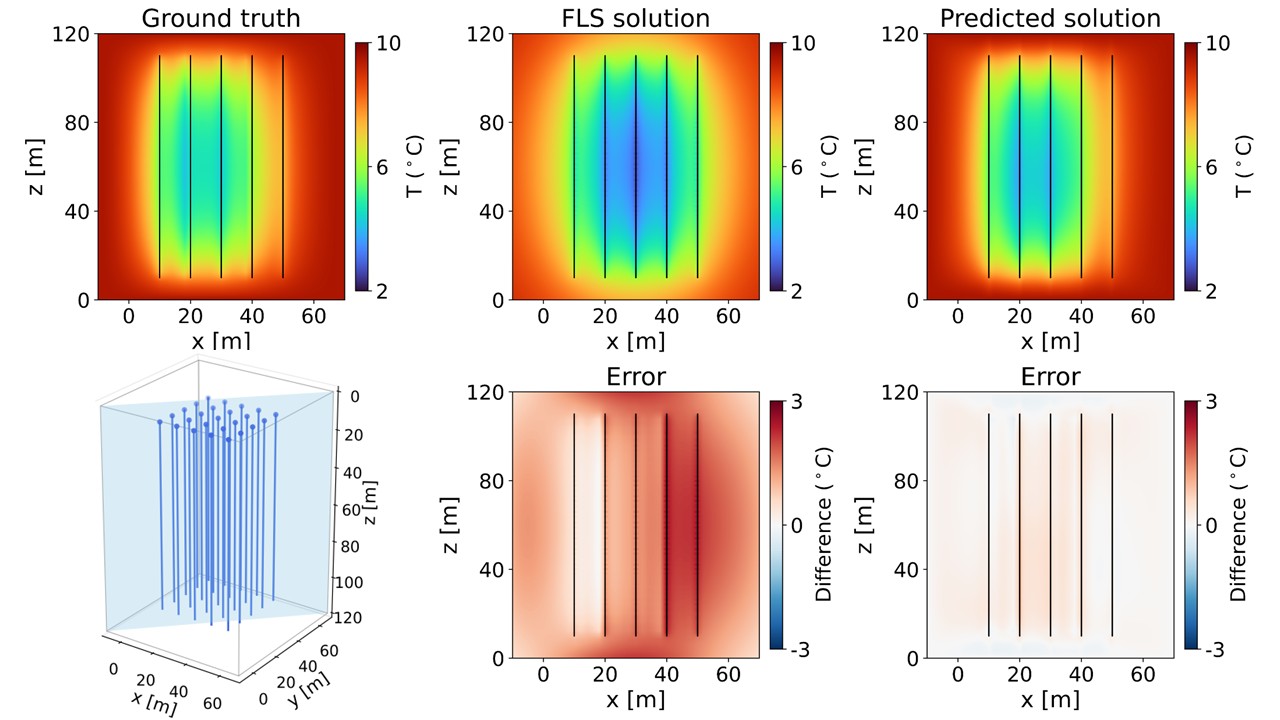}
    \caption{The first row: temperature distributions on the vertical slice at $y=x$ in December of year 30 for Scenario 2 under Dirichlet boundary condition, derived from the finite difference solver of mesh size $h=1$m (Ground truth), FLS analytical formula (FLS solution) and hybrid-PINN corrector (Predicted solution); the second row: the difference distributions of FLS solution and predicted solution with respect to the ground truth on the same cross section. The horizontal coordinate is labeled by the projection onto the $x$ axis.}
    \label{fig:fls_xy_30_error}
\end{figure}

\begin{table}[!htbp]
\centering
\caption{$L^2$ and $L^{\infty}$ relative errors of the global FLS solutions, local FLS solutions and hybrid-PINN predictions in Decembers of years $5$, $15$ and $30$ for Scenarios 1 and 2 under Dirichlet boundary conditions, computed over the vertical cross sections at $y=30$ m and $y=x$, respectively.}
\label{tab:fls_dirichlet_relative_errors}
\footnotesize
\setlength{\tabcolsep}{3.0pt}

\begin{tabular}{c c cc cc cc}
\toprule
\multirow{2}{*}{Scenario}
& \multirow{2}{*}{Month}
& \multicolumn{2}{c}{Global FLS}
& \multicolumn{2}{c}{Local FLS}
& \multicolumn{2}{c}{Predicted solution} \\
\cmidrule(lr){3-4}
\cmidrule(lr){5-6}
\cmidrule(lr){7-8}
&
& $L^2$ & $L^{\infty}$
& $L^2$ & $L^{\infty}$
& $L^2$ & $L^{\infty}$ \\
\midrule

\multirow{3}{*}{1}
& 60
& $3.9753\mathrm{e}{-2}$ & $1.1426\mathrm{e}{-1}$
& $2.8918\mathrm{e}{-2}$ & $1.0314\mathrm{e}{-1}$
& $1.3647\mathrm{e}{-2}$ & $6.5663\mathrm{e}{-2}$ \\

& 180
& $1.1527\mathrm{e}{-1}$ & $2.0087\mathrm{e}{-1}$
& $9.9794\mathrm{e}{-2}$ & $1.5969\mathrm{e}{-1}$
& $2.6352\mathrm{e}{-2}$ & $6.2691\mathrm{e}{-2}$ \\

& 360
& $2.0594\mathrm{e}{-1}$ & $3.0093\mathrm{e}{-1}$
& $2.0303\mathrm{e}{-1}$ & $2.7113\mathrm{e}{-1}$
& $3.3584\mathrm{e}{-2}$ & $8.4673\mathrm{e}{-2}$ \\

\midrule

\multirow{3}{*}{2}
& 60
& $3.5310\mathrm{e}{-2}$ & $1.3641\mathrm{e}{-1}$
& $2.7753\mathrm{e}{-2}$ & $1.1455\mathrm{e}{-1}$
& $1.2180\mathrm{e}{-2}$ & $6.6961\mathrm{e}{-2}$ \\

& 180
& $9.3774\mathrm{e}{-2}$ & $2.1277\mathrm{e}{-1}$
& $8.5121\mathrm{e}{-2}$ & $1.6648\mathrm{e}{-1}$
& $1.4805\mathrm{e}{-2}$ & $6.0220\mathrm{e}{-2}$ \\

& 360
& $1.7720\mathrm{e}{-1}$ & $3.1260\mathrm{e}{-1}$
& $1.7586\mathrm{e}{-1}$ & $2.6513\mathrm{e}{-1}$
& $2.4955\mathrm{e}{-2}$ & $8.7812\mathrm{e}{-2}$ \\

\bottomrule
\end{tabular}
\end{table}
Figures \ref{fig:ils_xy_30_error} and \ref{fig:fls_xy_30_error} illustrate the representative comparisons between the FDM-simulated, analytical and predicted temperature fields at the final time step for two test scenarios, showing better match of predicted solution to the reference one in both area with densely distributed BHEs and region near the boundary.

\paragraph{Supplementary results.} In the following, we provide supplementary results for the three examples in Section \ref{case_study} to further demonstrate the effectiveness of the proposed method. For brevity, we present the simulated solutions for Scenario 2 in the $5$th and $30$th years, together with the Scenario 1 MFLS results for the 15th year on the vertical cross sections at \(y=30\) m and \(x=30\) m.  
\begin{figure}[t]
    \centering
    \begin{subfigure}{\textwidth}
        \centering
        \includegraphics[width=\textwidth]{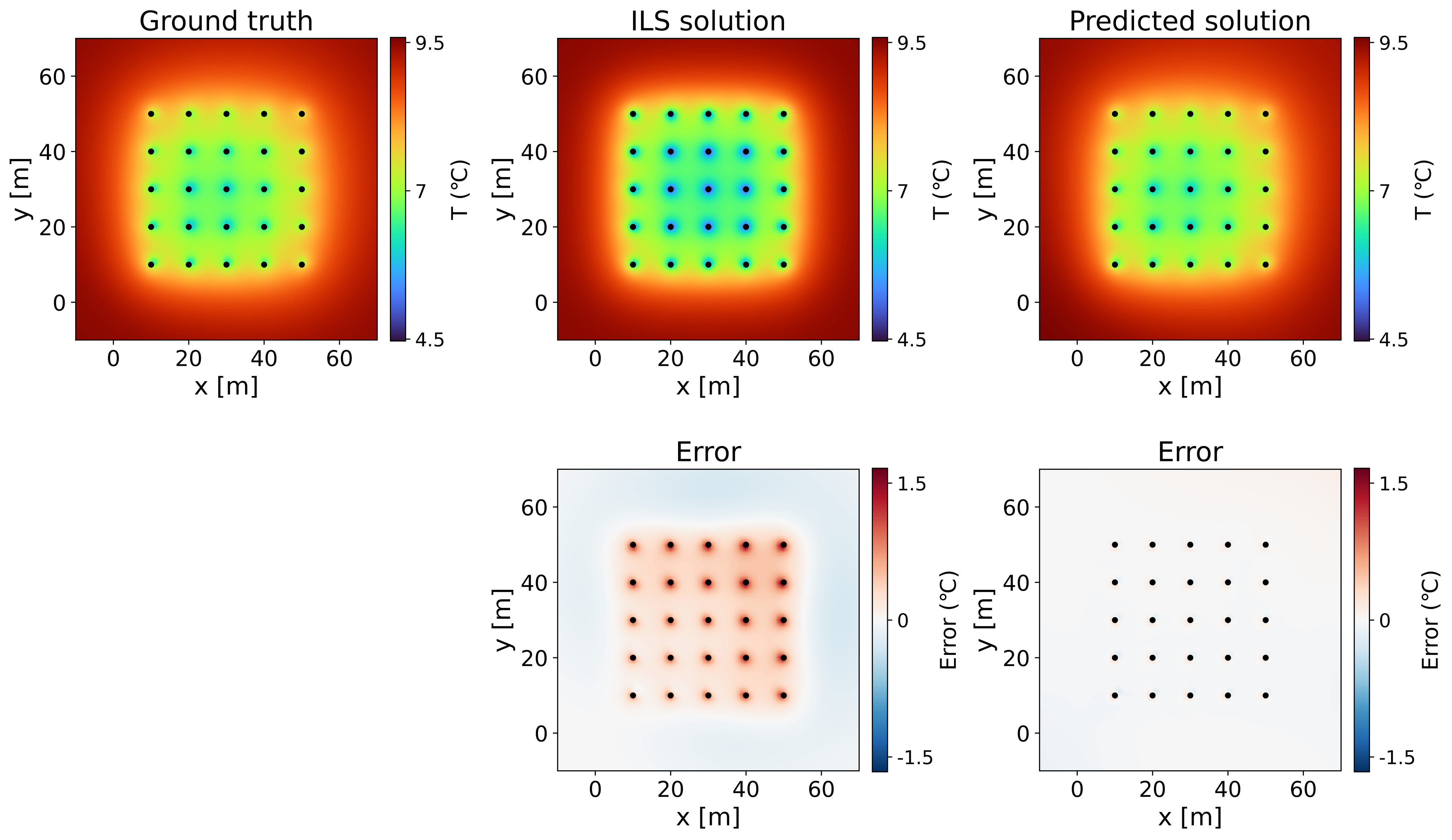}
        \caption{December of the $5$th simulated year.}
        \label{fig:ils_xy_5_open_error}
    \end{subfigure}
    
    \vspace{5mm} 
    
    \begin{subfigure}{\textwidth}
        \centering
        \includegraphics[width=\textwidth]{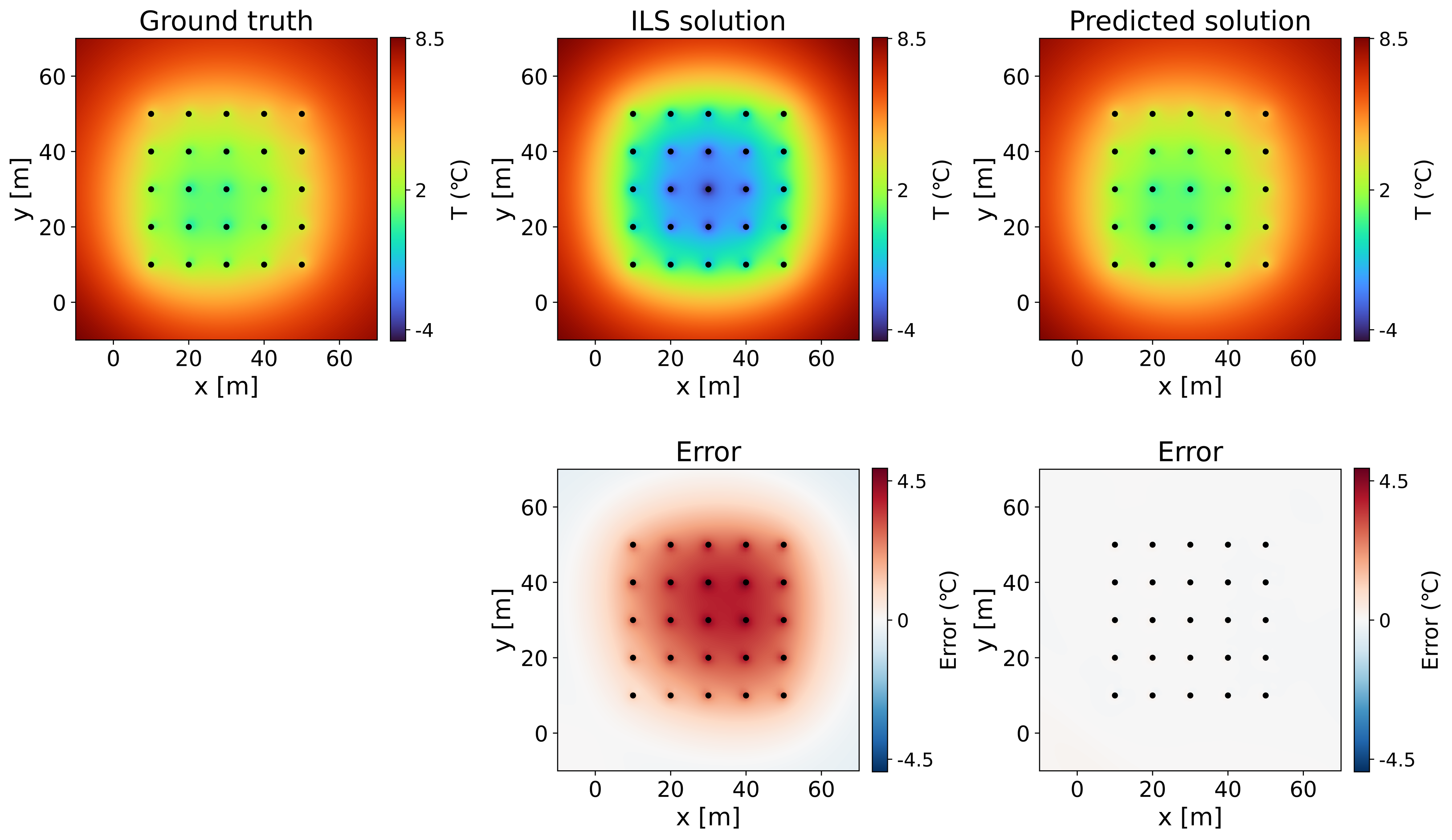}
        \caption{December of the $30$th simulated year.}
        \label{fig:ils_xy_30_open_error}
    \end{subfigure}
    \caption{Temperature distributions for Scenario 2, derived from the finite difference solver (Ground truth, $h=1$m), ILS analytical formula (ILS solution) and hybrid-PINN corrector (Predicted solution) and the difference distributions of ILS solution and predicted solution with respect to the ground truth.}
    \label{fig:ils_xy_linear_appendix}
\end{figure}

\begin{figure}[t]
    \centering
    \begin{subfigure}{\textwidth}
        \centering
        \includegraphics[width=\textwidth]{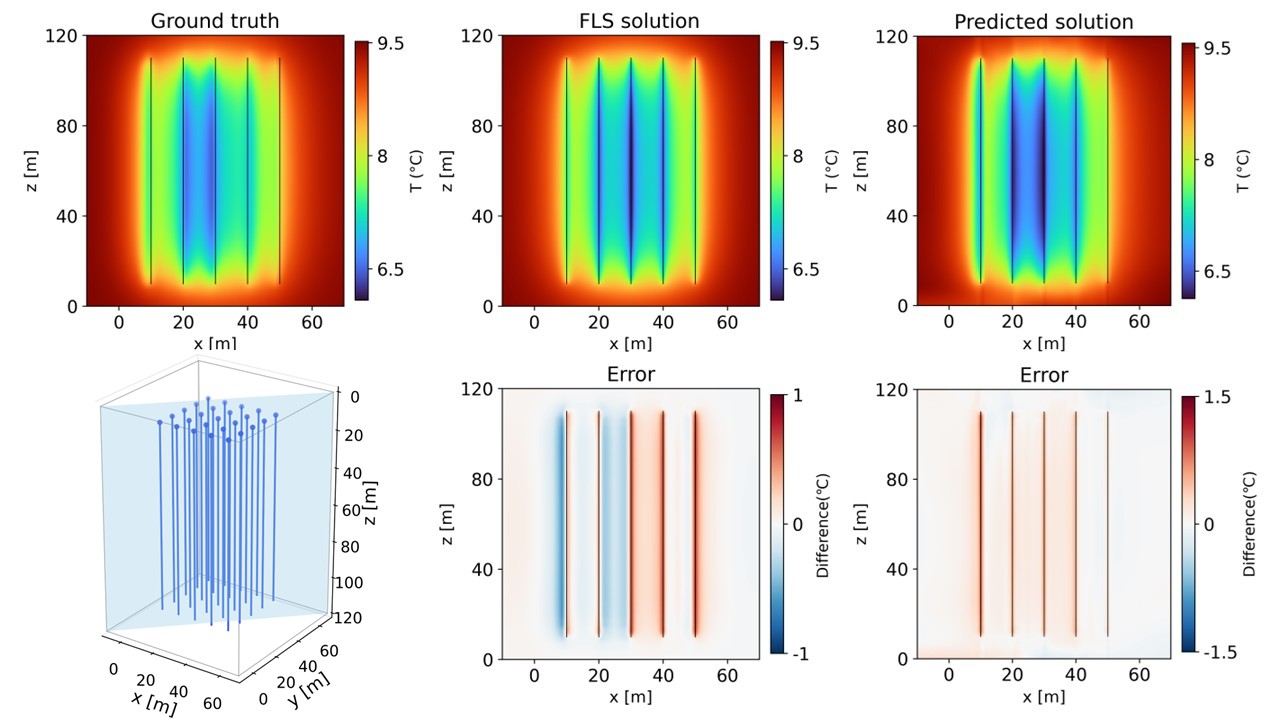}
        \caption{December of the $5$th simulated year.}
        \label{fig:fls_xy_5_open_error}
    \end{subfigure}
    
    \vspace{5mm} 
    
    \begin{subfigure}{\textwidth}
        \centering
        \includegraphics[width=\textwidth]{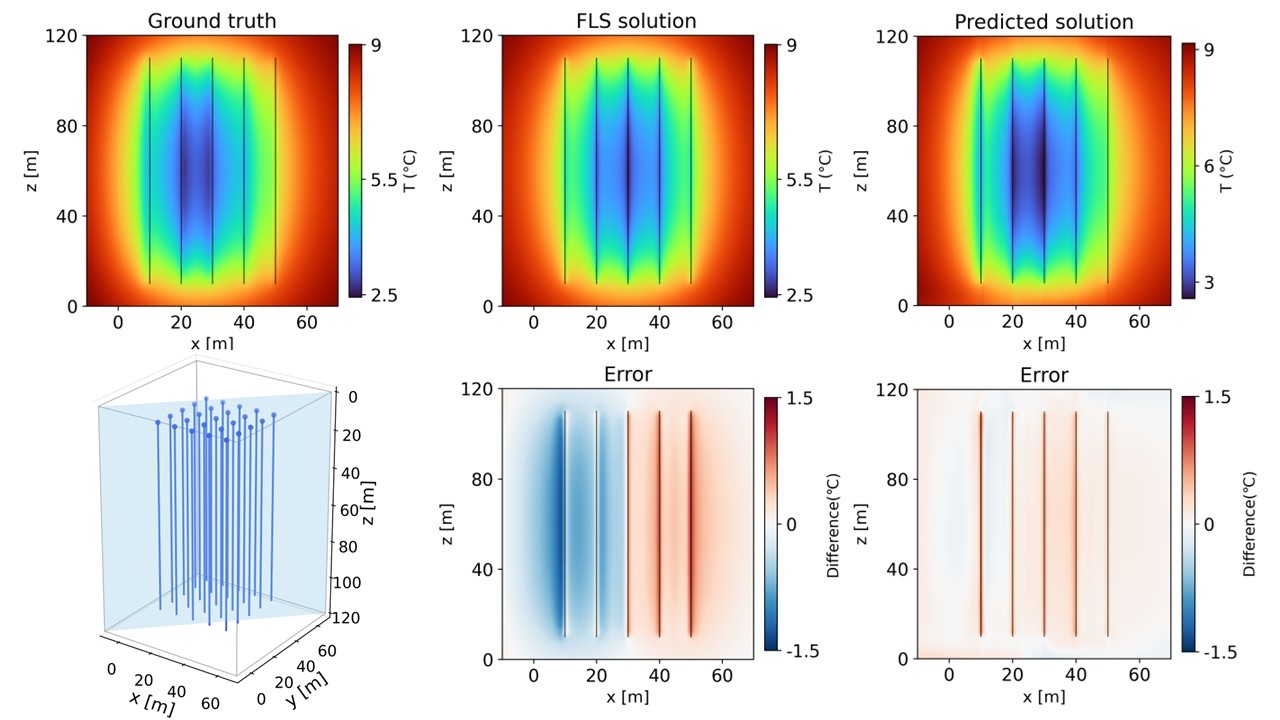}
        \caption{December of the $30$th simulated year.}
        \label{fig:fls_xy_30_open_error}
    \end{subfigure}
    \caption{Temperature distributions on the vertical section $y=x$ for Scenario 2, derived from the finite difference solver (Ground truth, $h=1$m), FLS analytical formula (FLS solution) and hybrid-PINN corrector (Predicted solution) and the difference distributions of FLS solution and predicted solution with respect to the ground truth on the same cross section. The horizontal coordinate is labeled by the projection onto the $x$ axis.}
    \label{fig:fls_xy_linear_appendix}
\end{figure}

\begin{figure}[t]
    \centering

    \begin{subfigure}{\textwidth}
        \centering
        \includegraphics[width=0.95\textwidth]{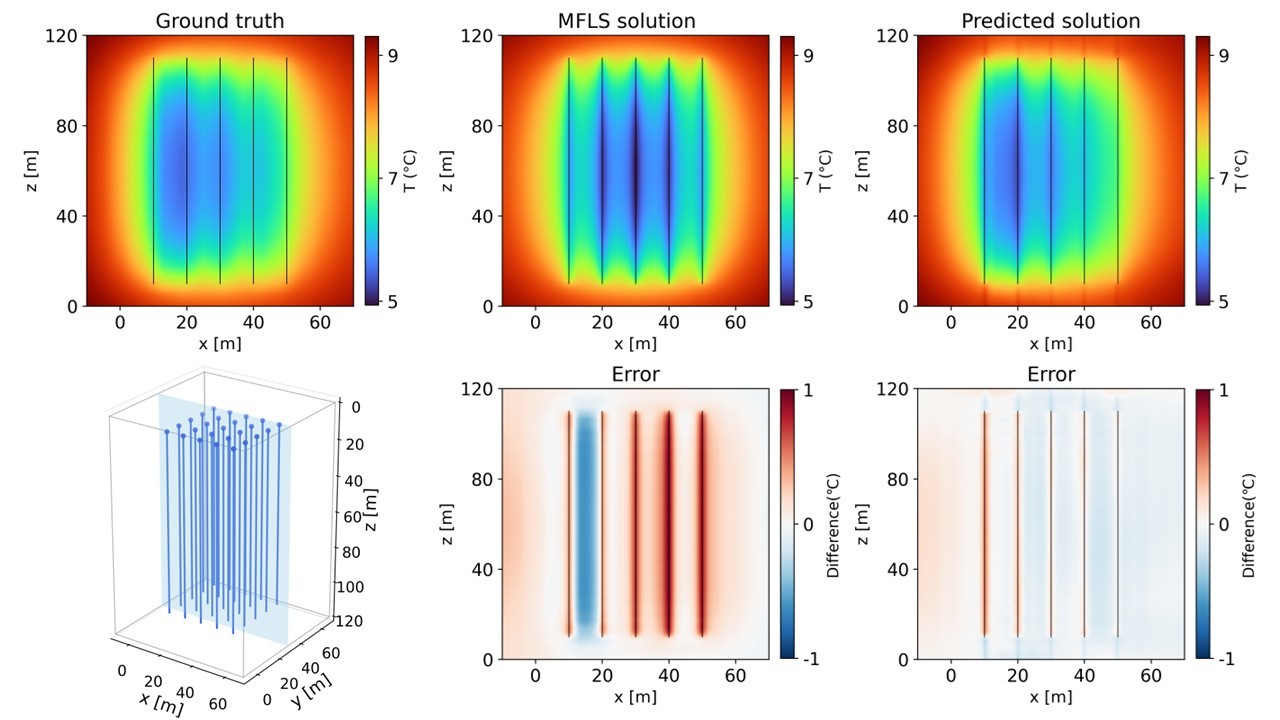}
        \caption{Vertical cross section at $y=30$ m.}
    \end{subfigure}

    \vspace{5mm}

    \begin{subfigure}{\textwidth}
        \centering
        \includegraphics[width=0.95\textwidth]{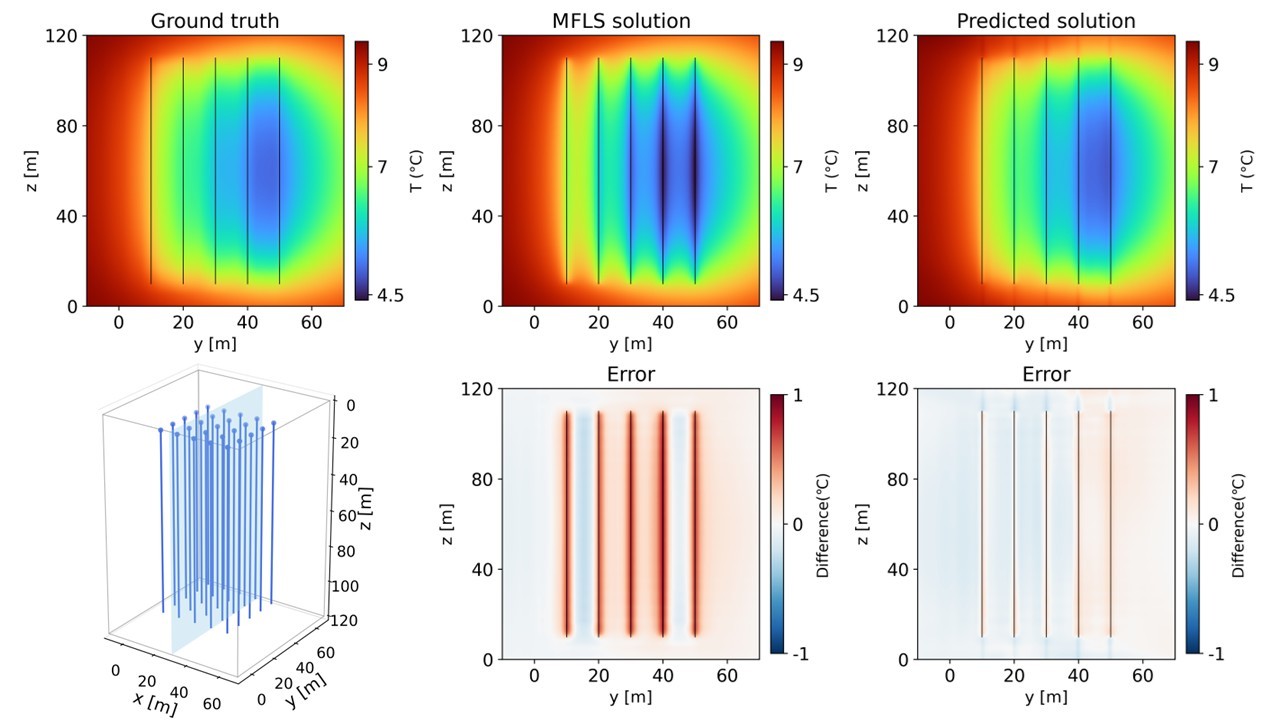}
        \caption{Vertical cross section at $x=30$ m.}
    \end{subfigure}

    \caption{Results in December of the $15$th year for Scenario 1 on (a) the vertical cross section at $y=30$ m and (b) the vertical cross section at $x=30$ m. For each panel: the first row, soil temperature distributions derived from the finite difference solver with mesh size $h=1$ m (Ground truth), global MFLS analytical formula (MFLS solution), and hybrid-PINN corrector based on the MFLS solution (Predicted solution); the second row, the difference distributions of the MFLS solution and predicted solution with respect to the ground truth.}
    \label{fig:mfls_x_15_error}
\end{figure}

\begin{figure}[t]
    \centering
    \begin{subfigure}{\textwidth}
        \centering
        \includegraphics[width=\textwidth]{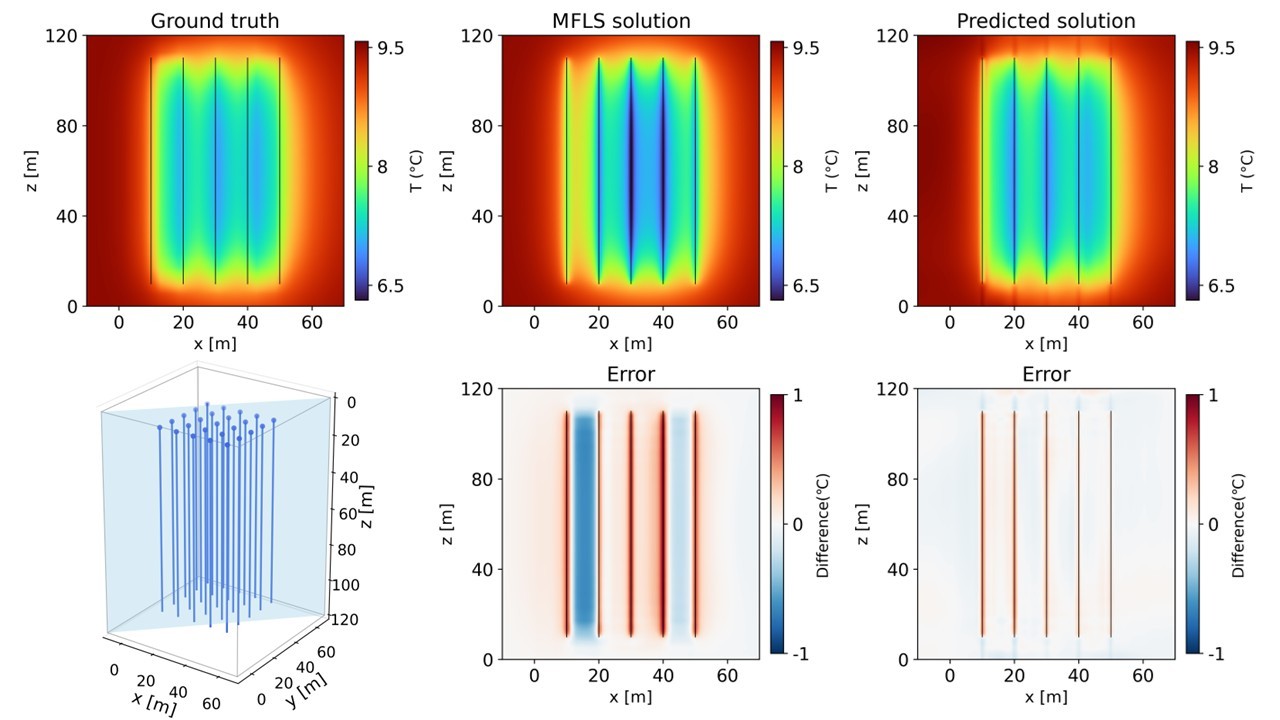}
        \caption{December of the $5$th simulated year.}
        \label{fig:mfls_xy_5_ymid_error}
    \end{subfigure}
    
    \vspace{5mm} 
    
    \begin{subfigure}{\textwidth}
        \centering
        \includegraphics[width=\textwidth]{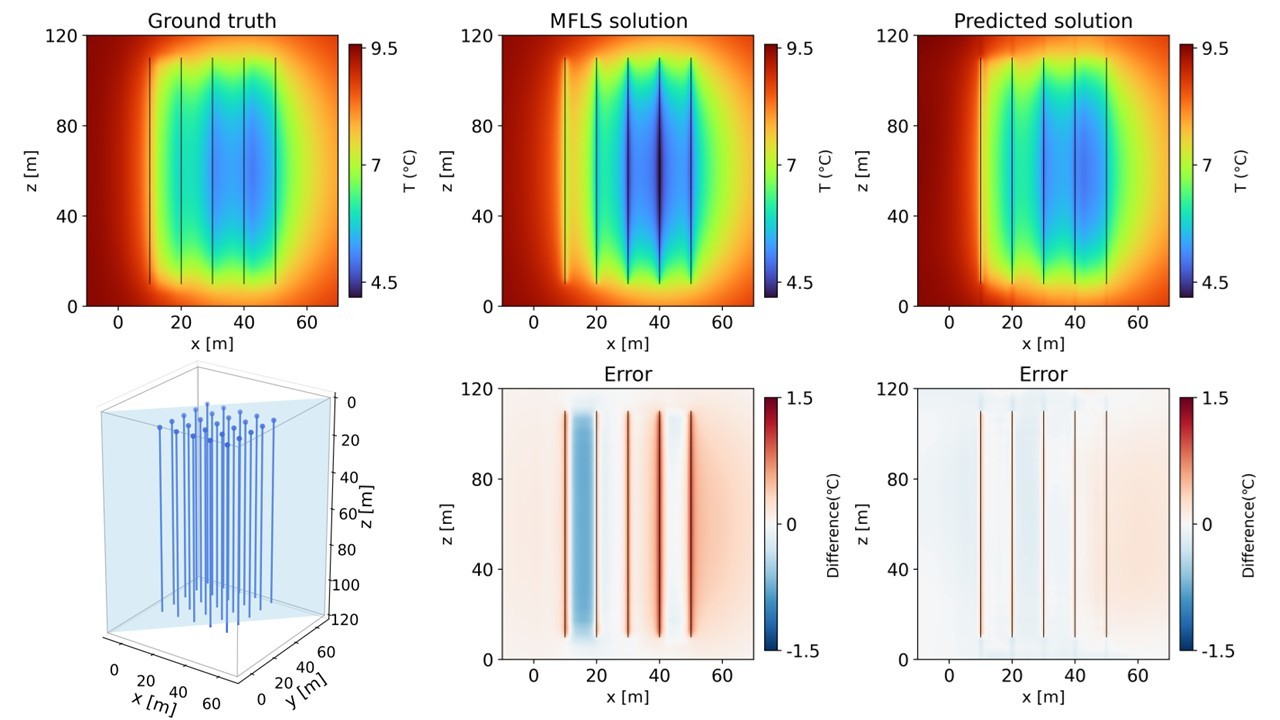}
        \caption{December of the $30$th simulated year.}
        \label{fig:mfls_xy_30_ymid_error}
    \end{subfigure}
    \caption{Temperature distributions on the vertical section $y=x$ for Scenario 2, derived from the finite difference solver (Ground truth, $h=1$m), MFLS analytical formula (MFLS solution) and hybrid-PINN corrector (Predicted solution) and the difference distributions of MFLS solution and predicted solution with respect to the ground truth on the same cross section. The horizontal coordinate is labeled by the projection onto the $x$ axis.}
    \label{fig:mfls_xy_linear_ymid_appendix}
\end{figure}

\begin{figure}[t]
    \centering
    \begin{subfigure}{\textwidth}
        \centering
        \includegraphics[width=\textwidth]{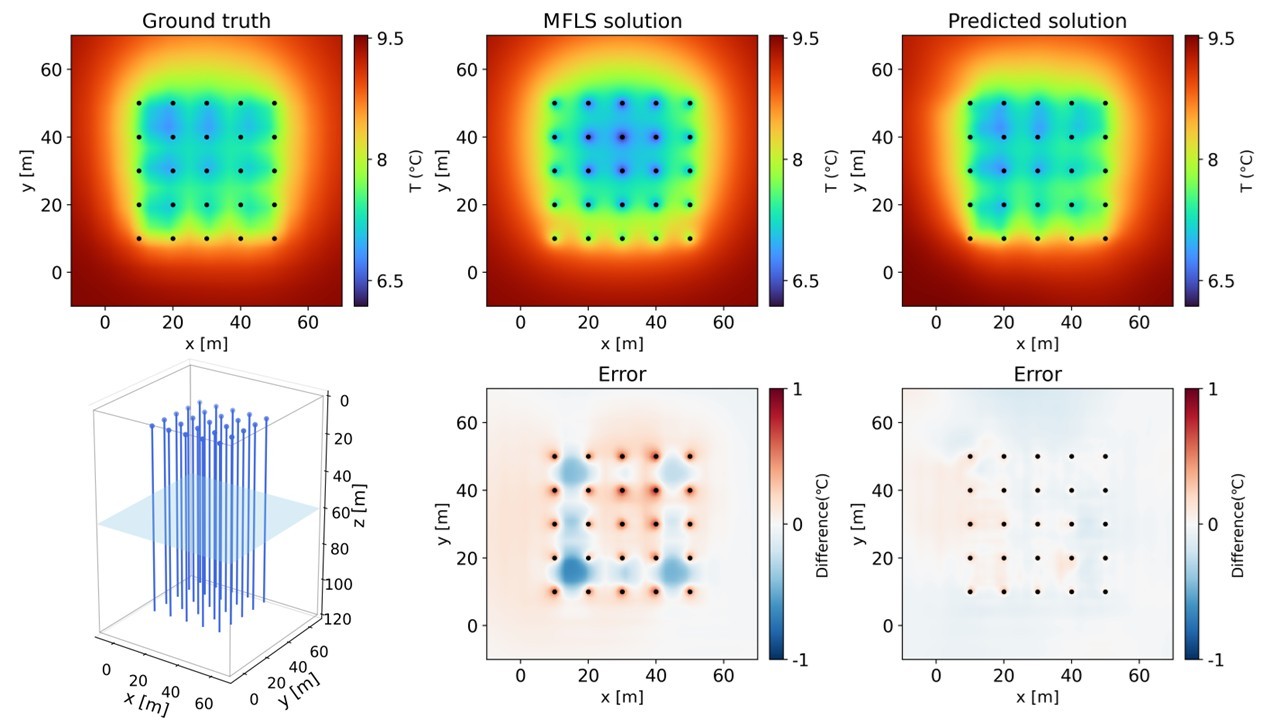}
        \caption{December of the $5$th simulated year.}
        \label{fig:mfls_xy_5_zmid_error}
    \end{subfigure}
    
    \vspace{5mm} 
    
    \begin{subfigure}{\textwidth}
        \centering
        \includegraphics[width=\textwidth]{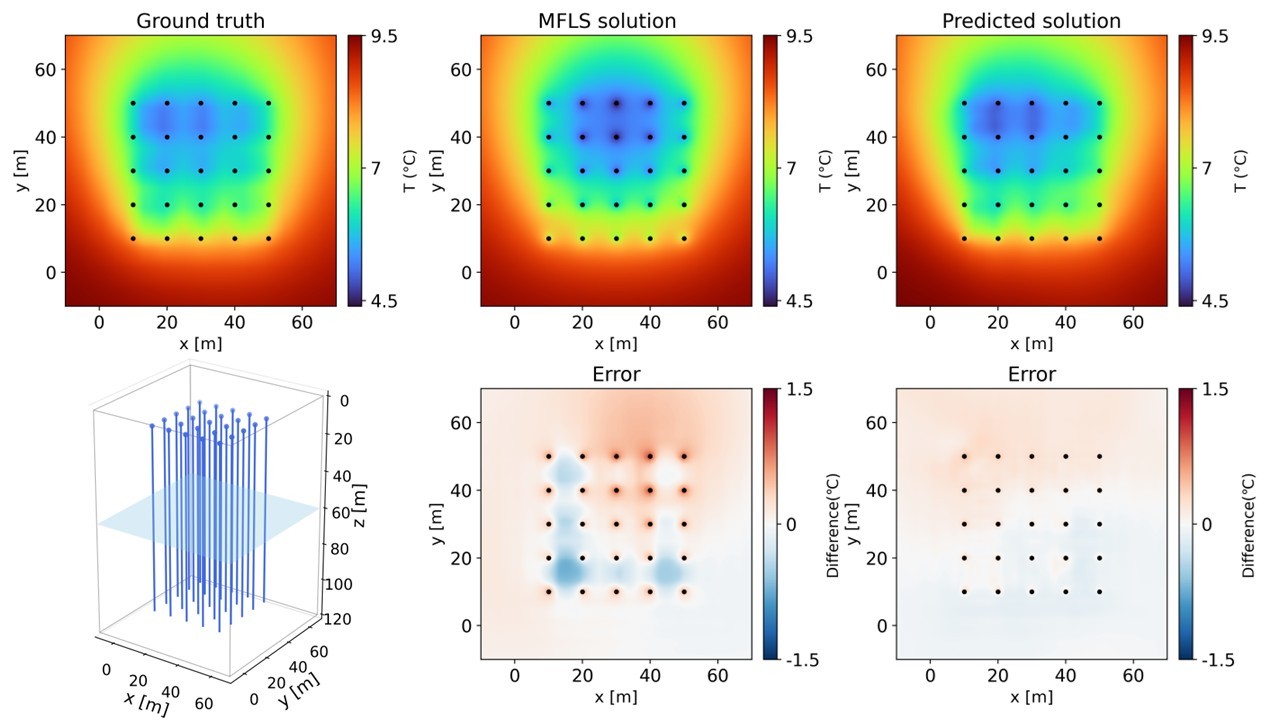}
        \caption{December of the $30$th simulated year.}
        \label{fig:mfls_xy_30_zmid_error}
    \end{subfigure}
    \caption{Temperature distributions on the horizontal section $z=60$ m for Scenario 2, derived from the finite difference solver (Ground truth, $h=1$m), MFLS analytical formula (MFLS solution) and hybrid-PINN corrector (Predicted solution) and the difference distributions of MFLS solution and predicted solution with respect to the ground truth on the same cross section.}
    \label{fig:mfls_xy_linear_zmid_appendix}
\end{figure}

\end{document}